\documentclass{article} 
\usepackage{iclr2027_conference,times}

\usepackage{amsmath,amsfonts,bm}

\def\eqref#1{equation~\ref{#1}}

\def\1{\bm{1}}

\DeclareMathAlphabet{\mathsfit}{\encodingdefault}{\sfdefault}{m}{sl}
\SetMathAlphabet{\mathsfit}{bold}{\encodingdefault}{\sfdefault}{bx}{n}

\def\gA{{\mathcal{A}}}

\def\gF{{\mathcal{F}}}

\def\gY{{\mathcal{Y}}}

\usepackage{hyperref}
\usepackage{url}
\usepackage{algorithm}
\usepackage{algpseudocode}
\usepackage{booktabs}
\usepackage{graphicx}
\usepackage{amsmath,amssymb, amsthm}
\usepackage{dsfont}
\usepackage{subcaption}
\usepackage{wrapfig}
\usepackage{mathtools}

\newtheorem{theorem}{Theorem}
\newtheorem{proposition}{Proposition}
\newtheorem{lemma}{Lemma}
\newtheorem{remark}{Remark}

\title{Sequential Anytime Valid Tests for Efficient Inference-Time Alignment}

\author{Vishesh Kumar Singh \& Sayak Ray Chowdhury\\
Indian Institute of Technology Kanpur, India\\
\texttt{\{visheshk24,sayakrc\}@iitk.ac.in} \\
}

\iclrfinalcopy 
\begin{document}

\maketitle

\begin{abstract}
Inference-time alignment methods, such as Best-of-N (BoN), improve language model outputs by generating many candidates per prompt and selecting among them using a reward model. The cost is largely wasted, since a fixed budget is applied uniformly across prompts, even though the computation required by different prompts varies widely. Adaptive methods halt generation once sampled answers appear settled, but their thresholds are tuned heuristics with no statistical control over early-stopping errors. Instead, we seek an anytime-valid stopping rule that takes a permitted Type-I error tolerance $\alpha$ as input, guaranteeing that generation halts only when an answer is certified to hold genuine majority dominance ($\theta_a > 1/2$) under the generator's distribution. To build such a rule, we aggregate candidates by their final answers using a reward-weighted self-consistency (RWSC) score and treat the decision of when to stop as a sequential test over the resulting clusters. Wald's sequential probability ratio test (SPRT) is the natural tool for such a test, and when applied to the RWSC clusters, it yields SPRT-RWSC, which performs well empirically. Yet its guarantee does not hold here, since the leading answer shifts as sampling proceeds, and the run is continuously monitored. Our main contribution replaces Wald's test with an anytime-valid construction that resolves this issue: it maintains a betting supermartingale (an e-process) for every answer the policy produces, bounding the probability of ever certifying a non-dominant answer by $\alpha$, uniformly over time and across all answers, however many the policy produces. We call the resulting rule e-RWSC: the RWSC scores drive the bets, which affect only how quickly a dominant answer $a^\star$ is certified, not the validity of the guarantee; and certification takes $O\bigl(\log(1/\alpha)/d_{\mathrm{KL}}(\theta_{a^\star}|| 1/2)\bigr)$ samples. Experiments across several language models, datasets, and reward models confirm the predicted $1/d_{\mathrm{KL}}$ scaling of per-response cost, and show that accuracy also improves over baselines at substantially lower token cost.
\end{abstract}

\section{Introduction}
\label{sec:intro}
Inference-time alignment has emerged as the practical alternative to fine-tuning, in which one repeatedly generates responses from the existing language model (e.g., a policy) and selects among them. Best-of-N \citep{beirami2024theoretical} selection scores each response with a reward model and returns the one with the highest score. Self-consistency~\citep{wang2023selfconsistency} takes a majority vote over final answers. Both are effective, but expensive: their cost is $N$ full
generations per prompt, and $N$ must be large enough for the hardest prompt in the workload. Since inference is the dominant cost of deployed language models, the token budget spent on selection should be minimized.

A fixed budget of $N$ is uniform across prompts, while the computation required by prompts varies enormously. On a prompt where the policy immediately concentrates on a single answer, ten samples are enough, whereas on a prompt on which the policy remains genuinely divided, even a thousand samples are not sufficient. Fixed-budget selection, therefore, results in both over- and underspending of tokens. Adaptive methods address this by halting generation once the sampled answers look settled, e.g., stopping when a posterior over answer frequencies concentrates~\citep{aggarwal2023lets, beacon2025}, when the marginal value of another sample
falls below its cost \citep{pandora2025}, or when a budget estimated from a short burn-in is exhausted \citep{adabon2025}. In every case, the halting rule is a tuned threshold with no control over the probability of stopping on the non-dominant answer. Moreover, they must be re-tuned for each new policy, reward model, and benchmark.

One would instead like an anytime-valid stopping rule that takes a formal error level $\alpha$ as input and adapts the token budget accordingly, guaranteeing that the probability of erroneously certifying a non-dominant answer ($\theta_{a^*} \le p^\dagger$) is strictly bounded by $\alpha$. The rule must be defined based on a computable statistic of the generations. Since many distinct reasoning traces can yield the same final answer, we group responses by their final answers and frame the stopping rule at the answer level. We introduce Reward-Weighted Self-Consistency (RWSC), which scores each cluster by the softmax-weighted mass of its members' reward scores, so that a cluster is favored both for containing many responses and for highly rewarded ones. Let $a^*$ denote the highest-scoring cluster (the leading answer) at the current step. Framing stopping as the question -- \emph{is $a^\star$ genuinely the dominant answer, or merely ahead by chance?} -- makes it a sequential hypothesis testing problem. Wald's sequential probability ratio test (SPRT) is the classic solution: as each new response arrives, one accumulates
evidence for or against the leader and stops once it crosses a boundary~\citep{wald1945sequential}. It performs strongly in practice, but remains a heuristic here, without any rigorous error-control guarantee.

Wald's guarantee assumes hypotheses that are fixed in advance, tested by a single terminal decision, and none of these holds here. Writing $\theta_a$ for the probability that a fresh response yields answer $a$,
the rates defining the test are estimated from the same responses the test then examines, rather than fixed beforehand. The leading answer $a^*$ is itself a function of the sampled responses, so the parameter under test, $\theta_{a^*}$, shifts whenever the leader does, forcing the test to restart. And the statistic is checked after every response, not at a single stopping point. Each of these
voids the error control that Wald's boundaries nominally provide.
We resolve this with anytime-valid inference
\citep{ramdas2023gametheoretic}, whose guarantees hold under continuous monitoring and arbitrary stopping. We maintain a separate betting supermartingale for each final answer the policy produces, instantiating one when that answer first appears and assigning it a weight from a fixed, summable sequence. Since the weights sum to 1, they split a single error budget across answers, regardless of how many arise. Ville's inequality~\citep{ville1939} then bounds by $\alpha$ the probability that a non-dominant answer ($\theta_a \le 1/2$) is ever certified, uniformly over time and over a growing set of answers. 
This addresses each failure of SPRT. A change in the leading answer is irrelevant, since every answer accumulates its own evidence. Continuous monitoring is an inherent property of Ville's inequality, so checking the boundary after each response incurs no additional cost.

Validity requires only that the bets be predictable, each computed from
earlier responses before the next one is sampled. The reward model sets each bet based on the current RWSC score of the answer cluster, which buys efficiency when the scores are accurate. The gain has a
precise form: under the log-optimal Kelly betting~\citep{kelly1956}, evidence for a dominant answer accrues at a rate
$d_{\mathrm{KL}}(\theta_a || 1/2)$, so certification takes about
$\log(1/\alpha) / d_{\mathrm{KL}}(\theta_a || 1/2)$ sampled responses. 
An accurate RWSC score bets near-optimally and approaches this rate, while an inaccurate or hacked score bets poorly and slows the certification. But it never crosses the boundary in expectation under the null, since validity does not depend on the estimate being correct.

The efficiency of the stopping rule, therefore, adapts to the prompt. The number of tokens spent scales with its difficulty: an answer the policy strongly favors ($\theta_a > 1/2$) is certified in a few responses, whereas one near the decision boundary $\theta_a = 1/2$ takes many, often more than the budget allows, in which case the procedure certifies nothing and falls back to standard aggregation. Either way, the practitioner fixes the error-control probability $\alpha$ once, and the number of responses adjusts itself to each prompt. We summarize our \emph{contributions} as follows.

\textbf{\textit{(i)}} We cast early stopping as a sequential hypothesis-testing problem and introduce \textsc{SPRT-RWSC}, which applies Wald's test to determine when the leading answer has converged. It operates on Reward-Weighted Self-Consistency (RWSC): an answer-level aggregation rule that scores answer clusters by reward-weighted mass, and recovers majority voting and Best-of-$N$ as extreme cases. It works well in practice, but Wald's error guarantee does not hold in this setting (Section~\ref{sec:sprt}).

\textbf{\textit{(ii)}} We propose \textsc{e-RWSC}, an anytime-valid stopping rule that accumulates evidence over all answers through separate e-processes. It controls the
Type-I error (certifying an answer that is not dominant under the policy) at a user-specified level $\alpha$, uniformly over time, and
over an unknown set of answers. The guarantee holds for any predictable choice of bets, so the reward model,
which sets them through RWSC scores, governs only efficiency and never validity. Under a plug-in estimate of the log-optimal Kelly betting, certifying the dominant answer (when one exists) takes
$O\bigl(\log(1/\alpha)/d_{\mathrm{KL}}(\theta_{a^*} || 1/2)\bigr)$ responses as $\alpha \to 0$. So, the number of tokens spent scales with the prompt's difficulty (Section~\ref{sec:e_rwsc}).

\textbf{\textit{(iii)}} An empirical study on 5 Language Models (Qwen2.5-3B, Qwen2.5-7B, Llama-3-8B, Phi-3-Mini-4K, Mistral-24B) with 4 Reward Models (ArmoRM, FsfairX, Skywork V1, Skywork V2) over 4 Benchmarks (GSM8K, MATH500, MMLU, CruxEval) showing that realized certification error stays below the nominal level, that per-response cost follows the predicted $1/d_{\mathrm{KL}}$ scaling with prompt difficulty, and that \textsc{e-RWSC} matches or outperform several baselines at substantially lower average cost (Section~\ref{sec:experiments}). For example, with Qwen-2.5-3B as the language model, ArmoRM as the reward model, and a maximum generation budget capped at 512, on the MATH500 Benchmark, we achieve an 8.2\% absolute accuracy gain over BoN with a 61.7\% smaller budget.

\textbf{Related work.} A line of work improves \emph{which} response is chosen from a pool: reward-weighted selection with a temperature~\citep{verdun2025soft}, ensemble reward models normalized across members~\citep{coste2023reward}, and step-level search or process verification that refine how each candidate is generated and scored~\citep{qiu2024treebon, chen2024alphamath,
chen2025sets, lee2025revise}. RWSC draws on the first two but aggregates over final answers rather than individual candidates. Our contribution concerns the orthogonal question of \emph{how many} candidates to generate. The adaptive stopping methods~\citep{aggarwal2023lets, beacon2025, pandora2025,
adabon2025, specrej2024} share a common limitation: their halting rules are tuned thresholds resting on distributional assumptions, with no control over the probability of stopping on the wrong answer. ConSol~\citep{consol2025}, like \textsc{SPRT-RWSC}, applies Wald's test to halt self-consistency generation, but compares only the top two answers and uses static priors. Two recent works give anytime-valid certificates for self-consistency:
MMC~\citep{cordero2025certified} certifies the empirical-leader mode via paired e-processes and couples this with test-time training, and CITE~\citep{ota2026cite} certifies a fixed target as the unique mode with a matching minimax lower bound. Unlike reward-agnostic mode certifiers (MMC, CITE), \textsc{e-RWSC} uses reward-informed betting to certify strict majority dominance ($p^\dagger \ge 0.5$), achieving $94.5\%$ accuracy on certified early stopped answers vs.\ $88.9\%$ for MMC with a lower budget.

\section{Answer-Level Aggregation and Sequential Stopping}
\label{sec:sprt}

Given a prompt $x$, a language model samples responses from a fixed reference
policy $\pi_{\mathrm{ref}}(\cdot | x)$. Each response $y$ carries a chain of
reasoning terminating in a final answer, extracted by a function
$\mathrm{Ans}(y)$ that maps $y$ to a canonical answer string (e.g., the final numerical result of a math problem). A reward model $r(x,y) \in \mathbb{R}$ scores how good a response is. \emph{Best-of-N} (BoN) samples $N$ responses and returns the one with the highest reward. \emph{Self-consistency} ignores the reward model and returns the answer produced by the majority of responses. Scoring responses individually, as BoN does, exposes selection to reward hacking at its worst: a single anomalously scored response wins outright. Yet discarding the reward signal entirely, as
self-consistency does, throws away information about response quality. Since many distinct reasoning chains reach the same final answer, we group responses at the answer level, scoring each answer group by both the number of responses it contains and the total reward it receives.

\subsection{Reward-Weighted Self-Consistency}
\label{sec:rwsc}

Let $\mathcal{Y}$ be a pool of $N$ sampled responses,
$\mathcal{Y}_a = \{ y \in \mathcal{Y} : \mathrm{Ans}(y) = a \}$ the set of responses yielding answer $a$, and $\mathcal{A} = \{ \mathrm{Ans}(y) : y \in \mathcal{Y} \}$ the set of observed
answers. Reward-Weighted Self-Consistency (RWSC) assigns each group the softmax-weighted mass of its members' rewards,
\begin{equation}
\label{eq:rwsc_score}
S(x, a) \;=\; \sum_{y \in \mathcal{Y}_a}
\frac{ \exp\bigl( \overline{r}(x, y) / \tau \bigr) }
     { \sum_{y' \in \mathcal{Y}} \exp\bigl( \overline{r}(x, y') / \tau \bigr) },
\qquad
\overline{r}(x, y) = \frac{ r(x, y) - \mu_x }{ \sigma_x },
\end{equation}
where $\overline{r}(x, y)$ is the reward normalized by the pool mean $\mu_x$ and standard deviation $\sigma_x$, and $\tau$ is a temperature parameter. The scores form a probability distribution over $\mathcal{A}$. As $\tau \to \infty$, every response contributes equally, and $S(x,a)$ reduces to the fraction of responses in cluster $a$, yielding self-consistency. As $\tau \to 0$, the mass
concentrates on the highest-reward response, recovering BoN. Intermediate $\tau$ allows the answer frequency and reward magnitude to jointly determine the winner. RWSC returns the answer with the highest mass and, within its cluster, the highest-reward response:
\begin{equation}
\label{eq:rwsc_ans}
a^* = \arg\max_{a \in \mathcal{A}} S(x, a),
\qquad
y^*_{\mathrm{RWSC}} = \arg\max_{y \in \mathcal{Y}_{a^*}} r(x, y).
\end{equation}
RWSC, like BoN and self-consistency, draws a fixed $N$ responses per prompt. This is wasteful, since the computation a prompt requires varies by orders of magnitude, and a budget large enough for the hardest prompt overspends on easy ones. Instead, we let each prompt govern its own budget, framing the decision of when to stop as a sequential test of the convergence of RWSC clusters.

\subsection{A Sequential Test over Answer Clusters}

Fix a prompt $x$. Let the responses $y_1, y_2, \dots$ be drawn sequentially from $\pi_{\mathrm{ref}}(\cdot | x)$. At each step $n$, let $\gY_n$ be the pool of sampled responses, $\gA_n$ the set of observed answers, $\gY_{n,a}$ the cluster for answer $a$
and $S_n(x,a)$ its RWSC mass. Let $a^*_{n} = \arg\max_{a \in \gA_{n}} S_n(x, a)$ be the current leader, the highest-RWSC-mass answer. For an answer $a$, write $\theta_a = \mathbb{P}_{y \sim \pi_{\mathrm{ref}}(\cdot \mid x)} \bigl( \mathrm{Ans}(y) = a \bigr)$
for its generation probability under $\pi_{\mathrm{ref}}$. Deciding when to stop amounts to determining whether the leading answer's probability of generation is sufficiently high to declare it dominant.  Each new response $y_{n+1}$ contributes to an agreement indicator
\begin{equation}
X_{n+1} = \mathds{1}\bigl\{ \mathrm{Ans}(y_{n+1}) = a^*_{n} \bigr\},
\quad \text{ with } \quad
\mathbb{P}\bigl( X_{n+1} = 1 \mid \mathcal{F}_{n-1} \bigr) =
\theta_{a^*_{n}} ,
\end{equation}
where $\gF_{n} = \sigma(\gY_n)$ denotes the sigma-algebra generated by first $n$ responses and their reward scores. The test asks whether the agreement rate $\theta_{a^*_{n}}$ is high, so that the leader $a^*_n$ has converged, or low, so that it has not. We instantiate it with Wald's sequential probability ratio test (SPRT) \citep{wald1945sequential} and call the resulting procedure \textsc{SPRT-RWSC}. Its pseudo-code is given in Appendix~\ref{app:code}.

\textbf{Hypothesis test}. Fix $0 < p_0 < 0.5 < p_1 < 1$ and at each step $n$, we test two \emph{simple}
hypotheses,
\begin{equation}
H_0 : \theta_{a^*_{n}} = p_0
\quad \text{against} \quad
H_1 : \theta_{a^*_{n}} = p_1 ,
\end{equation}
where $p_0$ represents a non-convergent regime in which the leader holds only a minority of the mass and $p_1$ a regime in which it is dominant. The two rates are set per prompt from a short burn-in of $n_0$ responses (e.g. $n_0 = 4$), during which clusters are formed but no evidence is accumulated: we set $p_0 = \min\{1/|\gA_{n_0}|, 0.5-\epsilon\}$ and $p_1 = \max\{S_{n_0}(x, a^*_{n_0}),\, 0.5+\epsilon\}$ for a small $\epsilon \in (0, 0.5)$, so that stopping always requires a strict majority. Under hypothesis $H_k$, $k \in \{0,1\}$, each observation has mass
$p_k^{X_i}(1 - p_k)^{1 - X_i}$, so the cumulative log-likelihood ratio after
$n$ responses,
\begin{equation}
\label{eq:llr}
\Lambda_n = \log \left(\prod_{i=n_0+1}^{n}
\frac{p_1^{X_i}(1-p_1)^{1-X_i}}{p_0^{X_i}(1-p_0)^{1-X_i}}\right)
= \sum_{i = n_0 + 1}^{n}
\left[ X_i \log \frac{p_1}{p_0}
+ (1 - X_i) \log \frac{1 - p_1}{1 - p_0} \right],
\end{equation}
accumulates evidence for $H_1$ over $H_0$. Given target error rates $\alpha$
(accepting a non-dominant leader) and $\beta$ (rejecting a dominant one) in $(0,1)$,
Wald's boundaries are $A = \log\frac{\beta}{1-\alpha}$ and
$B = \log\frac{1-\beta}{\alpha}$. If $\Lambda_n \ge B$, the test returns the current leader $a^*_n$ as the final answer. If $\Lambda_n \le A$, it rejects the leader and resets the evidence $\Lambda_n=0$. Otherwise, it draws another response, falling back to the current leader if the budget $N_{\max}$ is reached.

\textbf{Shifting leader.} The RWSC leader $a^*_n$ is a function of the responses and can change as we draw $y_{n+1}$. If the leader changes to $a^*_{n+1} \ne a^*_n$, the hypotheses -- defined relative to the previous leader -- no longer concern the quantity being tested, and the accumulated $\Lambda_n$ measures evidence about the wrong answer. \textsc{SPRT-RWSC} resets $\Lambda_n = 0$ and re-estimates $p_0, p_1$ for the new leader with updated answer set $\gA_{n+1}$ and leader's RWSC mass $S_{n+1}(x,a^*_{n+1})$. Thus, the hypotheses are repeatedly redefined based on the data, allowing the test to adapt to the prompt's difficulty.

However, one cannot control the errors here as one can with Wald's theorem. Wald's classical guarantee is for two \emph{known} hypotheses fixed in advance, with a single terminal decision. Here, the leader shifts during the run, changing the hypotheses, and the evidence $\Lambda_n$ is compared to the boundary after every response rather than at a single stopping point. \textsc{SPRT-RWSC} is thus a principled and effective heuristic, with tunable parameters $\alpha,\beta$, with no guarantee on Type I and II error.

\section{Anytime-Valid Certified Stopping}
\label{sec:e_rwsc}

Wald's guarantee is not designed for hypotheses that shift with the data or for a statistic inspected after every response. We therefore turn to a framework built for exactly these conditions. \emph{E-processes} \citep{ramdas2023gametheoretic}
retain their validity under optional stopping, and predictable data-dependent construction. By building the test on e-processes, we control the probability of ever certifying a non-dominant answer at a user-specified level, uniformly over time.

Given a prompt $x$, let $A_n = \mathrm{Ans}(y_n)$ be the final answer of response $y_n$ and $X_n^{(a)} = \mathds{1}\{A_n = a\}$ the indicator that response $n$ agrees with answer $a$. Let $\theta_a$ be the generation probability of answer $a$. Then, $X_n^{(a)}$, $n \ge 1$, are i.i.d.\ $\mathrm{Bernoulli}(\theta_a)$. For each answer $a$, we test 
\begin{equation}
H_0^{(a)}: \; \theta_a \le p^\dagger
\qquad \text{against} \qquad
H_1^{(a)}: \; \theta_a > p^\dagger,
\end{equation}
where $p^\dagger \in (0,1)$ is threshold controlling dominance of an answer. Under $H_0^{(a)}$, $a$ is not the dominant answer. Hence, rejecting $H_0^{(a)}$ certifies $a$ as dominant. When $p^\dagger \ge \tfrac12$, at most one answer can do so. So the certified answer, if it exists, is unique.

\textbf{Per-answer betting e-processes.}
Let $X_n^{(a)} = \mathds{1}\{A_n = a\}$ be the indicator that response $n$ agrees with answer $a$ and $\mathcal{F}_n^{(a)}=\sigma(y_1,X_1^{(a)},\ldots,y_n,X_n^{(a)})$ the sigma-algebra generated by the first $n$ responses, their reward scores, and agreement indicators with answer $a$. Let $(\lambda_n^{(a)})_{n \ge 1}$ be a stochastic process predictable with respect to the filtration $(\mathcal{F}_n^{(a)})_{n \ge 1}$, i.e., $\lambda_n^{(a)}$ is $\mathcal{F}_{n-1}^{(a)}$-measurable.
Let $\sigma_a = \min\{ n : A_n = a \}$ the first appearance of answer $a$. Define, for $n \ge \sigma_a$, the stochastic process
\begin{equation}
\label{eq:eprocess}
E_n^{(a)} = \prod\nolimits_{i = \sigma_a + 1}^{n}
\Bigl( 1 + \lambda_i^{(a)} \bigl( X_i^{(a)} - p^\dagger \bigr) \Bigr),
\quad E_{\sigma_a}^{(a)} = 1.
\end{equation}
The above admits a betting interpretation~\citep{ramdas2023gametheoretic}: a bettor, starting with unit capital, bets repeatedly on the event that the next candidate agrees with the answer $a$, at odds set by the null $\theta_a \le p^\dagger$. Staking a fraction $\lambda_i^{(a)} p^\dagger$ of current capital returns $\lambda_i^{(a)}(1 - p^\dagger)$ on agreement and
forfeits $\lambda_i^{(a)} p^\dagger$ otherwise, so $E_n^{(a)}$ is the bettor's wealth after $n$ rounds. 

\textbf{Stopping rule.} Let $J(a) \in \{1, 2, \dots\}$ be the rank of answer $a$ in order of first appearance. Assign weights $\pi_j = \tfrac{1}{j(j+1)}$, so that $\sum_{j \ge 1} \pi_j = 1$.
Given a certification level $\alpha \in (0,1)$, we stop generating responses at
\begin{equation}
\label{eq:stopping_time}
N = \min\bigl\{ n \ge 1 : \exists\, a \text{ with }
\pi_{J(a)}\, E_n^{(a)} \ge \tfrac{1}{\alpha} \bigr\} \wedge N_{\max}.
\end{equation}
If some answer $a^*$ triggers the boundary, it is \emph{certified}, and the returned candidate is the highest-reward trajectory in its cluster, $y^* = \arg\max_{y \in \mathcal{Y}_{a^*}} r(x, y)$. If the budget $N_{\max}$ is exhausted first, the procedure falls back to uncertified RWSC selection (\eqref{eq:rwsc_ans}). 
\begin{theorem}[Anytime-valid certification]
\label{thm:av}
Fix $\alpha, p^\dagger,c \in (0,1)$. For any answer distribution $\{\theta_a\}_a$, any budget $N_{\max} \in \mathbb{N} \cup \{\infty\}$, and any predictable sequence of bets $(\lambda_i^{(a)})_{i,a} \subset [0,\, c/p^\dagger]$,
\begin{equation}
\mathbb{P}\left( \exists\, n,\; \exists\, a \text{ with }
\theta_a \le p^\dagger : a \text{ is certified at time } n \right)
\le \alpha .
\end{equation}
Equivalently, with probability greater than $1 - \alpha$, uniformly over all $n$ and over all answer $a$ such that $a$ is certified at time $n$, satisfies $\theta_a > p^\dagger$. 
\end{theorem}
\emph{Proof sketch.} Certification of answer $a$ occurs when the bettor's wealth reaches $1/(\pi_{J(a)}\alpha)$. For an answer that is not dominant ($\theta_a \le p^\dagger$), $E_n^{(a)}$ is a nonnegative supermartingale, so by Ville's inequality, it attains this level with probability at most $\pi_{J(a)}\alpha$. Summing the weights $\pi_{J(a)}$ over answers bounds the overall probability of ever certifying a non-dominant answer by $\alpha$. Complete proof is given in Appendix~\ref{app:proof}.
\begin{remark}
The bets enter Theorem~\ref{thm:av} only through their range and predictability; they may depend arbitrarily on past responses and rewards, so the reward model cannot affect the certificate's validity. A change in the RWSC leader is likewise immaterial: each answer accumulates evidence in its own e-process, so no statistic is invalidated or discarded when the leader shifts. Finally, Theorem~\ref{thm:av} certifies \emph{dominance under the policy}, $\theta_{a^*} > p^\dagger$, not correctness: a confidently mistaken policy
is certified quickly on a wrong answer. What links certification to accuracy is the self-consistency premise that the majority vote is usually correct, which we
verify empirically (Section~\ref{sec:experiments}).
\end{remark}

\subsection{Reward-informed betting}

We now turn to that efficiency: given the freedom to choose any predictable bet, which choice certifies the dominant answer fastest, and how does the reward model inform that choice? Since the wealth $E_n^{(a)}$ compounds multiplicatively, the natural efficiency criterion is the \emph{growth rate} $\tfrac{1}{n} \log E_n^{(a)}$, and the bet maximizing it is the classical Kelly bet~\citep{kelly1956, breiman1961}.
\begin{lemma}[Kelly bet]
\label{lem:kelly}
Let $X \sim \mathrm{Bernoulli}(\theta)$ and $\lambda \in [0, c/p^\dagger]$ for $c \in (0,1)$.
Let
$G_\theta(\lambda) = \mathbb{E}[\log(1 + \lambda(X - p^\dagger))]$
be the expected one-step log-increment in wealth. Then $G_\theta$ is strictly concave,
and its unconstrained maximizer is the \emph{Kelly bet} $\lambda^\star(\theta) = \frac{\theta - p^\dagger}{p^\dagger(1 - p^\dagger)}$. When admissible, i.e. $\lambda^\star(\theta) \le c/p^\dagger$, it attains the growth rate $G_\theta(\lambda^\star(\theta))=d_{\mathrm{KL}}(\theta \,\|\, p^\dagger)$.
\end{lemma}
Applied to answer $a$, the numerator of the bet is the bettor's edge over null, and the denominator is the variance of evidence under null. So the optimal stake is proportional to the signal-to-noise ratio of the evidence, and it is non-positive exactly when $\theta_a \le p^\dagger$, where no favorable wager exists. Maximizing expected log wealth, rather than expected wealth, is necessary: expected wealth is maximized by the all-in bet $\lambda = 1/p^\dagger$, which collapses the wealth $E_n^{(a)}$ to zero on the first disagreement and cannot recover. Restricting to $\lambda \le c/p^\dagger$ with $c < 1$ rules this out: every step then multiplies the wealth
by at least $1 - c > 0$, so a disagreement shrinks $E_n^{(a)}$ but never collapses it, and the Kelly bet
$\lambda^\star(\theta_a)$ within this range maximizes the growth rate $G_{\theta_a}(\lambda)$. When the bet is
admissible, its growth rate is the KL-divergence between true and null agreement rates: evidence accrues at a rate equal to the statistical distinguishability of the two hypotheses. It matches the classical efficiency of Wald's SPRT~\citep{wald1945sequential} while retaining validity under composite nulls and optional stopping.

\textbf{Plug-in bets.} The Kelly bet depends on the unknown generation probability $\theta_a$ and so cannot be played. Instead, we form a predictable estimate of $\theta_a$ from the responses and their rewards observed so far. At each step $n\ge \sigma_a +1$, where $\sigma_a$ is the first appearance of answer $a$, we play the bet
\begin{equation}
\label{eq:plugin_bet}
\lambda_n^{(a)} =
\mathrm{clip}\left(
\frac{\widehat{\theta}_{n-1}^{(a)} - p^\dagger}{p^\dagger (1 - p^\dagger)},
\; 0, \; \frac{c}{p^\dagger} \right),
\quad \text{where} \quad
\widehat{\theta}_{n-1}^{(a)} =
\frac{ Z_{n-1}^{(a)} + m \, S_{n-1}(x, a) }
     { (n - 1 - \sigma_a) + m }.
\end{equation}
Here $Z_{n-1}^{(a)}=\sum_{j = \sigma_a + 1}^{n-1} X_j^{(a)}$ denotes the number of agreements it accrued since its appearance, and hence the empirical agreement rate $Z_{n-1}^{(a)}/(n-1-\sigma_a)$ serves as an estimate of $\theta_a$. We use a regularized estimator, where the current RWSC mass $S_{n-1}(x, a)$ of its cluster (\eqref{eq:rwsc_score}) acts as a prior pseudo-observation of strength $m$ (we use $m = 4$). The reward model enters the bet and, consequently, the test through the RWSC mass. We call this procedure \textsc{e-RWSC}. Its pseudo-code is given in Appendix~\ref{app:code}.

The clip keeps the bet in the admissible range $[0,\, c/p^\dagger]$. Informative reward scores steer $\widehat\theta_n^{(a)}$ toward $\theta_a$
early, sharpening the bets and accelerating certification; miscalibrated scores merely slow the growth of capital and cost efficiency, without affecting validity, since Theorem~\ref{thm:av} holds for \emph{any} predictable
bet. Proposition~\ref{prop:growth} makes this precise, giving the asymptotic growth rate of the capital and the resulting certification time under the plug-in bet.
\begin{proposition}[Growth rate and certification time]
\label{prop:growth}
Let $N_{\max} \in \mathbb{N} \cup \lbrace\infty\rbrace$, $\theta_{a^*} > p^\dagger$ for an answer $a^\star$ and the Kelly bet $\lambda^\star(\theta_{a^\star}) \le c/p^\dagger$ for $c \in (0,1)$.
Then, as $n\to \infty$, the plug-in bet $\lambda_n^{(a^\star)} \to \lambda^\star(\theta_{a^\star})$, and the capital $\tfrac{1}{n} \log E_n^{(a^\star)} \to
d_{\mathrm{KL}}\bigl(\theta_{a^\star} || p^\dagger\bigr)$, almost surely. Let $N_{a^\star}(\alpha) = \min\{ n : \pi_{J(a^\star)}
E_n^{(a^\star)} \ge 1/\alpha \}$ be the time at which $a^\star$ is certified. Then $N_{a^\star}(\alpha) < \infty$ and as $\alpha \to 0$, $\tfrac{N_{a^\star}(\alpha)}{\log(1/\alpha)}
\to \frac{1}{d_{\mathrm{KL}}(\theta_{a^\star} || p^\dagger)}$ almost surely. Moreover, when $p^\dagger \ge \tfrac12$, the stopping time $N$ of~\eqref{eq:stopping_time} coincides with $N_{a^\star}(\alpha)$ except on an event of probability at most $\alpha$. consequently
$N/\log(1/\alpha) \to 1/d_{\mathrm{KL}}(\theta_{a^\star} || p^\dagger)$ in probability as $\alpha \to 0$.
\end{proposition}
Proposition~\ref{prop:growth} says that \textsc{e-RWSC} certifies a dominant answer in about $\log(1/\alpha)/d_{\mathrm{KL}}(\theta_{a^\star} || p^\dagger)$ responses: the cost grows only logarithmically in the confidence $1/\alpha$ and scales inversely with
$d_{\mathrm{KL}}(\theta_{a^\star} \,\|\, p^\dagger)$, which measures how far the answer's mass sits above the threshold $p^\dagger$. Easy prompts, where one answer clearly dominates, are certified in a handful of responses; near-ties cost many, so computational cost adapts automatically to prompt difficulty. This rate is essentially the best attainable. 
\citet{ota2026cite} proves a lower bound of the same order for sequential \emph{mode} certification under self-consistency, where the sample complexity is governed by the margin between the target answer and its closest competitor. Instead, \textsc{e-RWSC} certifies that the returned answer holds an absolute majority, $\theta_a > \tfrac12$ -- stronger than being the mode -- with governing quantity
$d_{\mathrm{KL}}(\theta_a || \tfrac12)$. The rate transfers to the stopping time $N$ of the test (\eqref{eq:stopping_time}) in probability.
Proof is in Appendix~\ref{app:proof}.
\begin{remark}
\label{rem:concurrent}
\citet{ota2026cite} certify self-consistency by an intersection-union test (\textsc{Cite}), with a confidence-weighted variant in which each vote carries a weight. Here, the weight enters the certified object, so a poor weight changes what is certified. In \textsc{e-RWSC}, the reward model enters only the bets, leaving the certified event
$\theta_a > \tfrac12$ fixed, so a poor reward can only slow
certification, never invalidate it. More importantly, \textsc{Cite} fixes the target in advance, whereas \textsc{e-RWSC} runs per-answer e-processes and certifies whichever answer crosses first, uniformly over the growing answer set. Empirically, \textsc{e-RWSC} achieves higher accuracy at a lower average token cost than \textsc{Cite}.
\end{remark}

\section{Experiments}
\label{sec:experiments}
To evaluate the efficacy and generalizability of our proposed E-SPRT framework, we conduct experiments across four datasets. We utilize GSM8K~\citep{cobbe2021gsm8k} and MATH500~\citep{hendrycks2021measuring} to assess multi-step mathematical reasoning. To evaluate broad domain knowledge and academic reasoning, we benchmark on MMLU~\citep{hendrycks2021mmlu}. We test the framework's capability in code reasoning and execution tracking using the CruxEval benchmark~\citep{gu2024cruxeval}. The generation capabilities of our framework were analyzed across four distinct base language models: Qwen2.5-3B, Qwen2.5-7B, Llama-3-8B, and Phi-3-Mini-4K. To evaluate the generated reasoning paths, we utilized a diverse ensemble of four reward models: ArmoRM~\citep{armorm}, Skywork V1~\citep{skywork_rm} and V2~\citep{skywork-v2}, and FsfairX~\citep{fsfairx}.

\begin{figure}[t]
\centering
\begin{subfigure}{\textwidth}
    \centering
    \includegraphics[width=\textwidth]{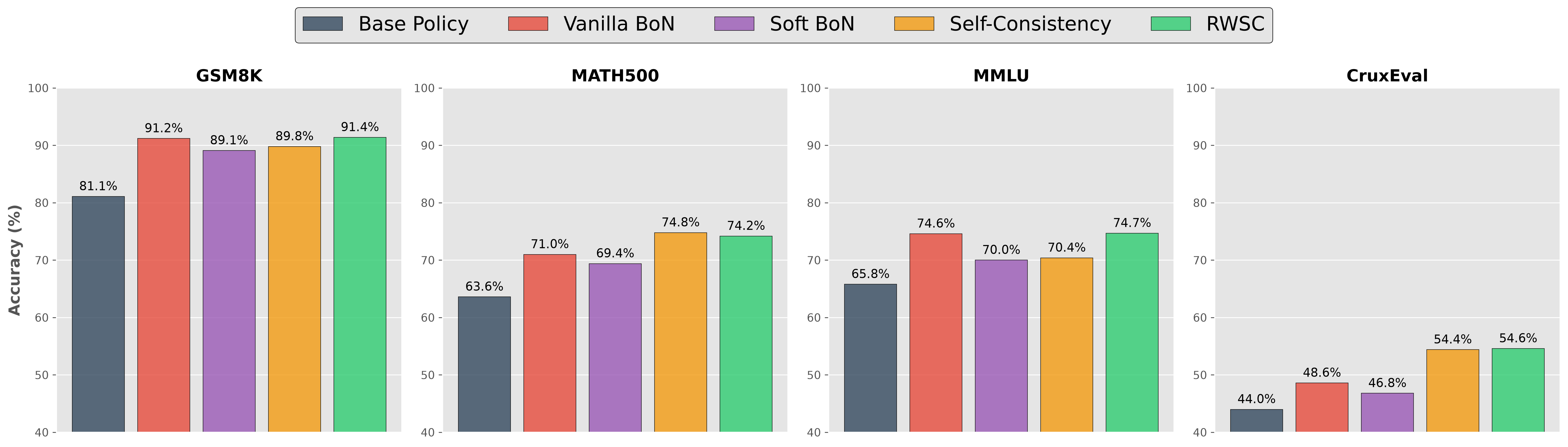}
    \caption{Full Budget Baselines Accuracy}
\end{subfigure}
\vspace{0.1cm}
\begin{subfigure}{\textwidth}
    \centering
    \includegraphics[width=\textwidth]{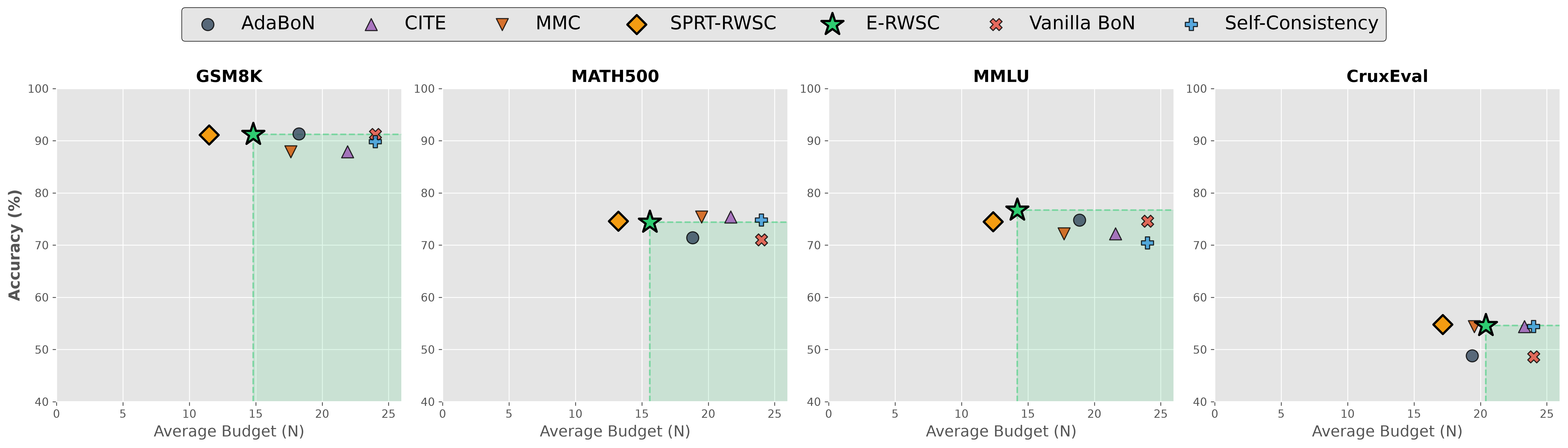}
    \caption{Early-stopping Pareto Frontiers (Accuracy vs. Average Budget)}
\end{subfigure}
\vspace{-2.5em}
\caption{\footnotesize{Comprehensive comparison of all selection methods across GSM8K, MATH500, MMLU and CruxEval for \textbf{Qwen2.5-3B-Instruct} using \textbf{ArmoRM} reward model, with allowed maximum budget as 24. For E-Process methods, we use $\alpha=0.1$, $p^\dagger=0.5$, and $c=0.8$. For SPRT we use $\alpha{=}0.1, \beta{=}0.2$.}}
\vspace{-1.5em}
\label{fig:qwen_armorm}
\end{figure}

\textbf{Comparison with Baselines.} Figure \ref{fig:qwen_armorm} (a) compares RWSC against full-budget baselines and Figure \ref{fig:qwen_armorm} (b) compares our framework SPRT-RWSC and E-RWSC against early-stopping baselines with $N_{\max}=24$. Across all evaluated tasks, RWSC consistently outperforms all full-budget baselines, achieving up to a 6\% accuracy improvement over Vanilla BoN on CruxEval. Among sequential methods, AdaBoN~\citep{adabon2025} maintains baseline accuracy but fails to meaningfully reduce compute, consistently requiring ${\sim}18{-}19$ candidates on average. In contrast, SPRT-RWSC and E-RWSC successfully cut the required budget nearly in half (${\sim}11{-}17$ candidates) while simultaneously matching or strictly exceeding the accuracy of full-budget baselines. As highlighted by the shaded dominance regions in Figure \ref{fig:qwen_armorm} (b), our methods, E-RWSC and SPRT-RWSC, consistently achieve the optimal Pareto frontier, outperforming CITE on both accuracy and budget reduction over 3 benchmarks and demonstrating a superior balance between alignment quality and computational efficiency. For a more detailed comparison with all the baselines, refer to Table \ref{tab:qwen_armorm} in the appendix.

\textbf{Cross-Model Generalization and Reward Model Sensitivity.}
We conduct a two-dimensional ablation on MATH500. Table~\ref{tab:cross_model_math500} applies a fixed reward model (ArmoRM) across four generators (3B to 8B). While absolute performance scales with base policy strength (peaking at 81.8\% for Qwen2.5-7B), SPRT-RWSC and E-RWSC consistently dominate BoN accuracy across all models, cutting inference budgets by up to 54\% (less on the harder-to-converge Llama-3-8B and Phi-3-Mini-4K). Moreover, the stopping rule adapts to generator confidence, expanding budgets for uncertain models (22.45 for Llama-3-8B) and halting early for highly convergent ones (13.06 for Qwen2.5-7B). Table \ref{tab:cross_rm_math500} fixes the generator (Qwen2.5-3B) to test RM sensitivity. BoN is sensitive to RM calibration, dropping to 62.6\% with the weaker FsfairX RM and surging to 76.8\% with Skywork V2. In contrast, RWSC and its sequential variants act as powerful regularizers against RM failures. E-RWSC maintains a robust 72.6\% accuracy even with FsfairX and scales to 78.6\% with Skywork V2.

\begin{table}[t]
\centering
\resizebox{\columnwidth}{!}{%
\begin{tabular}{l | cc | cc | cc | cc | cc}
\toprule
& \multicolumn{2}{c|}{\textbf{Qwen2.5-3B}} & \multicolumn{2}{c|}{\textbf{Qwen2.5-7B}} & \multicolumn{2}{c|}{\textbf{Llama-3-8B}} & \multicolumn{2}{c|}{\textbf{Phi-3-Mini-4K}} & \multicolumn{2}{c}{\textbf{Mistral-24B}} \\
\cmidrule(lr){2-3}\cmidrule(lr){4-5}\cmidrule(lr){6-7}\cmidrule(lr){8-9}\cmidrule(lr){10-11}
\textbf{Method} & \textbf{Acc} & \textbf{Budget} & \textbf{Acc} & \textbf{Budget} & \textbf{Acc} & \textbf{Budget} & \textbf{Acc} & \textbf{Budget} & \textbf{Acc} & \textbf{Budget} \\
\midrule
$\pi_{\text{ref}}$ (Base) & 63.6\% & 1.00 & 75.6\% & 1.00 & 26.2\% & 1.00 & 39.2\% & 1.00 & 66.2\% & 1.00 \\
Vanilla BoN     & 71.0\% & 24.00 & 77.8\% & 24.00 & 32.6\% & 24.00 & 48.2\% & 24.00 & 72.0\% & 24.00 \\
RWSC            & 74.2\% & 24.00 & 81.4\% & 24.00 & 39.6\% & 24.00 & 56.4\% & 24.00 & 76.8\% & 24.00 \\
\midrule
Pandora ($c=0.05$) & 68.6\% & $4.00_{\pm 0.00}$ & 78.0\% & $4.00_{\pm 0.00}$ & 30.0\% & $4.00_{\pm 0.00}$ & 44.2\% & $4.00_{\pm 0.00}$ & 70.8\% & $4.00_{\pm 0.00}$ \\
AdaBoN           & 71.4\% & $18.82_{\pm 5.20}$ & 78.0\% & $18.97_{\pm 5.26}$ & 32.2\% & $18.90_{\pm 5.31}$ & 47.8\% & $18.84_{\pm 5.48}$ & 70.8\% & $18.67_{\pm 5.79}$ \\
BEACON           & 70.8\% & $22.55_{\pm 4.27}$ & 77.6\% & $22.28_{\pm 4.70}$ & 32.0\% & $22.39_{\pm 4.57}$ & 47.8\% & $22.66_{\pm 4.12}$ & 72.2\% & $22.09_{\pm 4.98}$ \\
ConSol           & 75.4\% & $8.02_{\pm 7.38}$ & 80.8\% & $5.91_{\pm 5.99}$ & 38.6\% & $14.69_{\pm 8.38}$ & 53.4\% & $12.02_{\pm 8.64}$ & 76.8\% & $7.62_{\pm 7.40}$ \\
CITE ($\alpha=0.1$) & 75.4\% & $21.68_{ \pm 2.85 }$ & 81.8\% & $20.78_{ \pm 2.92 }$ & 39.0\% & $23.74_{ \pm 1.15 }$ & 53.6\% & $23.11_{ \pm 2.02 }$ & 77.4\% & $21.45_{ \pm 2.88 }$ \\
MMC ($\alpha=0.1$)  & 75.4\% & $19.50_{\pm 7.49}$ & 81.6\% & $19.94_{\pm 7.16}$ & 39.0\% & $21.68_{\pm 5.61}$ & 53.6\% & $20.22_{\pm 6.95}$ & 77.4\% & $19.24_{\pm 7.64}$ \\
\midrule
\multicolumn{11}{l}{\textit{Ours (Proposed)}} \\
SPRT-RWSC ($n_0=4$)  & 74.6\% & $13.23_{\pm 7.30}$ & 81.6\% & $11.09_{\pm 6.46}$ & 39.6\% & $19.33_{\pm 6.28}$ & 56.6\% & $16.70_{\pm 7.26}$ & 76.8\% & $12.87_{\pm 7.22}$ \\
E-RWSC ($\alpha=0.1$) & 74.4\% & $15.59_{\pm 8.20}$ & 81.4\% & $13.06_{\pm 7.81}$ & 39.6\% & $22.45_{\pm 4.63}$ & 56.4\% & $20.16_{\pm 6.75}$ & 76.8\% & $15.02_{\pm 8.11}$ \\
\bottomrule
\end{tabular}
}
\caption{Cross-Model Performance Comparison on MATH500 ($N=24$, ArmoRM).}
\label{tab:cross_model_math500}
\end{table}

\begin{table}[t]
\centering
\vspace*{-2mm}
\resizebox{\columnwidth}{!}{%
\begin{tabular}{l | cc | cc | cc | cc}
\toprule
& \multicolumn{2}{c|}{\textbf{ArmoRM}} & \multicolumn{2}{c|}{\textbf{Skywork V1}} & \multicolumn{2}{c|}{\textbf{FsfairX}} & \multicolumn{2}{c}{\textbf{Skywork V2}} \\
\cmidrule(lr){2-3}\cmidrule(lr){4-5}\cmidrule(lr){6-7}\cmidrule(lr){8-9}
\textbf{Method} & \textbf{Acc} & \textbf{Budget} & \textbf{Acc} & \textbf{Budget} & \textbf{Acc} & \textbf{Budget} & \textbf{Acc} & \textbf{Budget} \\
\midrule
$\pi_{\text{ref}}$ (Base) & 63.6\% & 1.00 & 63.6\% & 1.00 & 63.6\% & 1.00 & 63.6\% & 1.00 \\
Vanilla BoN     & 71.0\% & 24.00 & 68.6\% & 24.00 & 62.6\% & 24.00 & 76.8\% & 24.00 \\
RWSC            & 74.2\% & 24.00 & 76.0\% & 24.00 & 72.8\% & 24.00 & 78.8\% & 24.00 \\
\midrule
Pandora ($c=0.05$) & 68.6\% & $4.00_{\pm 0.00}$ & 69.0\% & $14.91_{\pm 8.27}$ & 64.0\% & $6.11_{\pm 5.08}$ & 72.4\% & $9.73_{\pm 7.74}$ \\
AdaBoN           & 71.4\% & $18.82_{\pm 5.20}$ & 69.0\% & $19.59_{\pm 5.14}$ & 63.0\% & $19.09_{\pm 5.29}$ & 75.8\% & $19.10_{\pm 5.56}$ \\
BEACON           & 70.8\% & $22.55_{\pm 4.27}$ & 69.4\% & $21.53_{\pm 5.65}$ & 62.6\% & $22.23_{\pm 4.91}$ & 77.2\% & $21.77_{\pm 5.41}$ \\
ConSol           & 75.4\% & $8.02_{\pm 7.38}$ & 75.4\% & $8.02_{\pm 7.38}$ & 75.4\% & $8.02_{\pm 7.38}$ & 75.4\% & $8.02_{\pm 7.38}$ \\
CITE ($\alpha=0.1$) & 75.4\% & $21.68_{ \pm 2.85 }$ & 75.4\% & $21.68_{ \pm 2.85 }$ & 75.4\% & $21.68_{ \pm 2.85 }$ & 75.4\% & $21.68_{ \pm 2.85 }$ \\
MMC ($\alpha=0.1$)  & 75.4\% & $19.50_{\pm 7.49}$ & 75.4\% & $19.50_{\pm 7.49}$ & 75.4\% & $19.50_{\pm 7.49}$ & 75.4\% & $19.50_{\pm 7.49}$ \\
\midrule
\multicolumn{9}{l}{\textit{Ours (Proposed)}} \\
SPRT-RWSC ($n_0=4$)  & 74.6\% & $13.23_{\pm 7.30}$ & 75.8\% & $13.61_{\pm 7.54}$ & 72.6\% & $13.51_{\pm 7.39}$ & 78.2\% & $12.94_{\pm 7.17}$ \\
E-RWSC ($\alpha=0.1$) & 74.4\% & $15.59_{\pm 8.20}$ & 76.0\% & $15.63_{\pm 8.19}$ & 72.6\% & $15.65_{\pm 8.23}$ & 78.6\% & $15.50_{\pm 8.18}$ \\
\bottomrule
\end{tabular}
}
\caption{Cross-RM Sensitivity Analysis on MATH500 (Qwen2.5-3B, $N=24$).}
\label{tab:cross_rm_math500}
\end{table}

\begin{figure}[ht]
    \centering
    \begin{subfigure}[b]{0.53\linewidth}
        \centering
        \includegraphics[width=\linewidth]{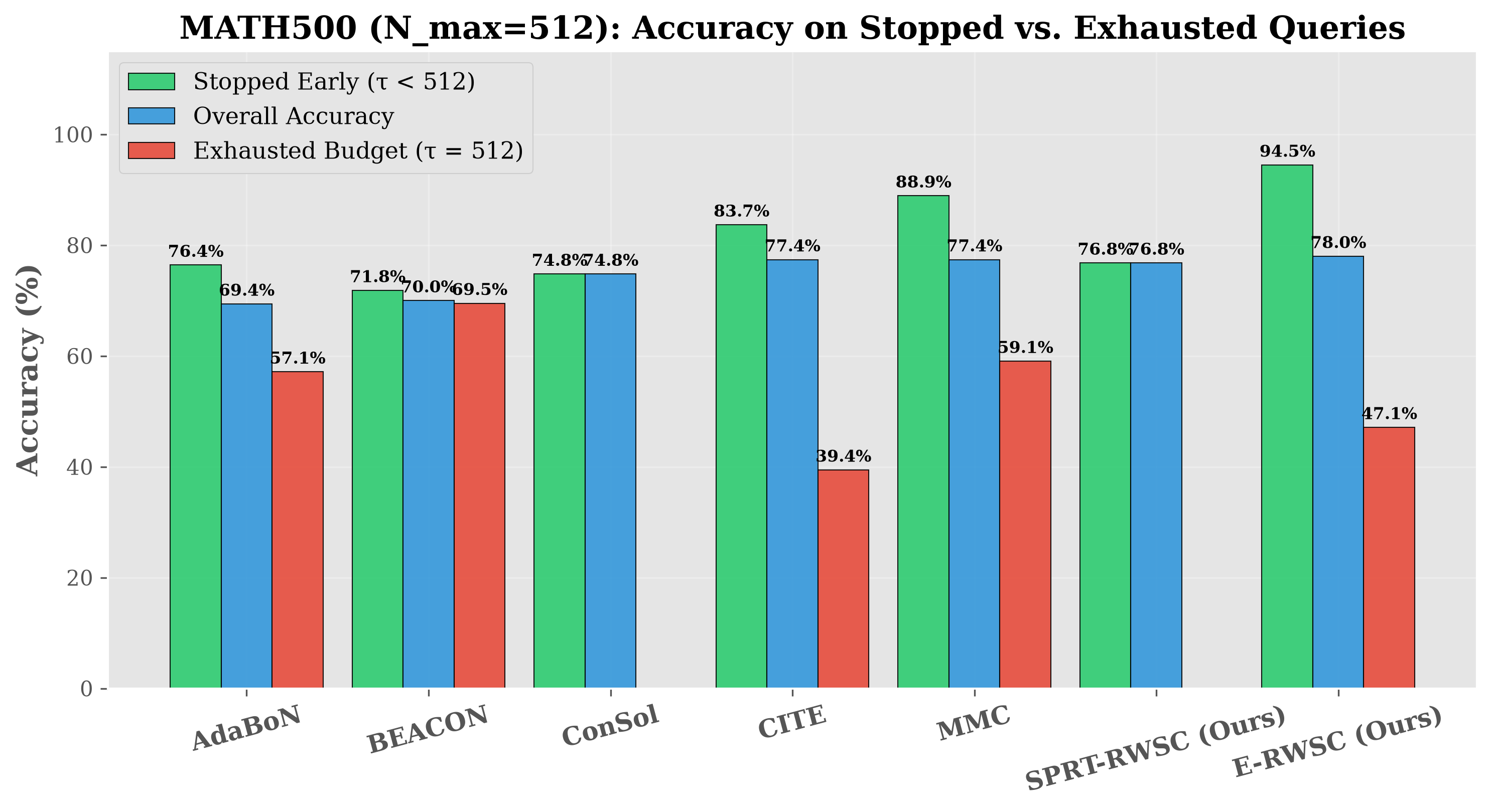}
        \caption{Accuracy on stopped ($\tau < 512$) vs. exhausted queries.}
        \label{fig:stopped_vs_exhausted_acc}
    \end{subfigure}
    \hfill
    \begin{subfigure}[b]{0.45\linewidth}
        \centering
        \includegraphics[width=\linewidth]{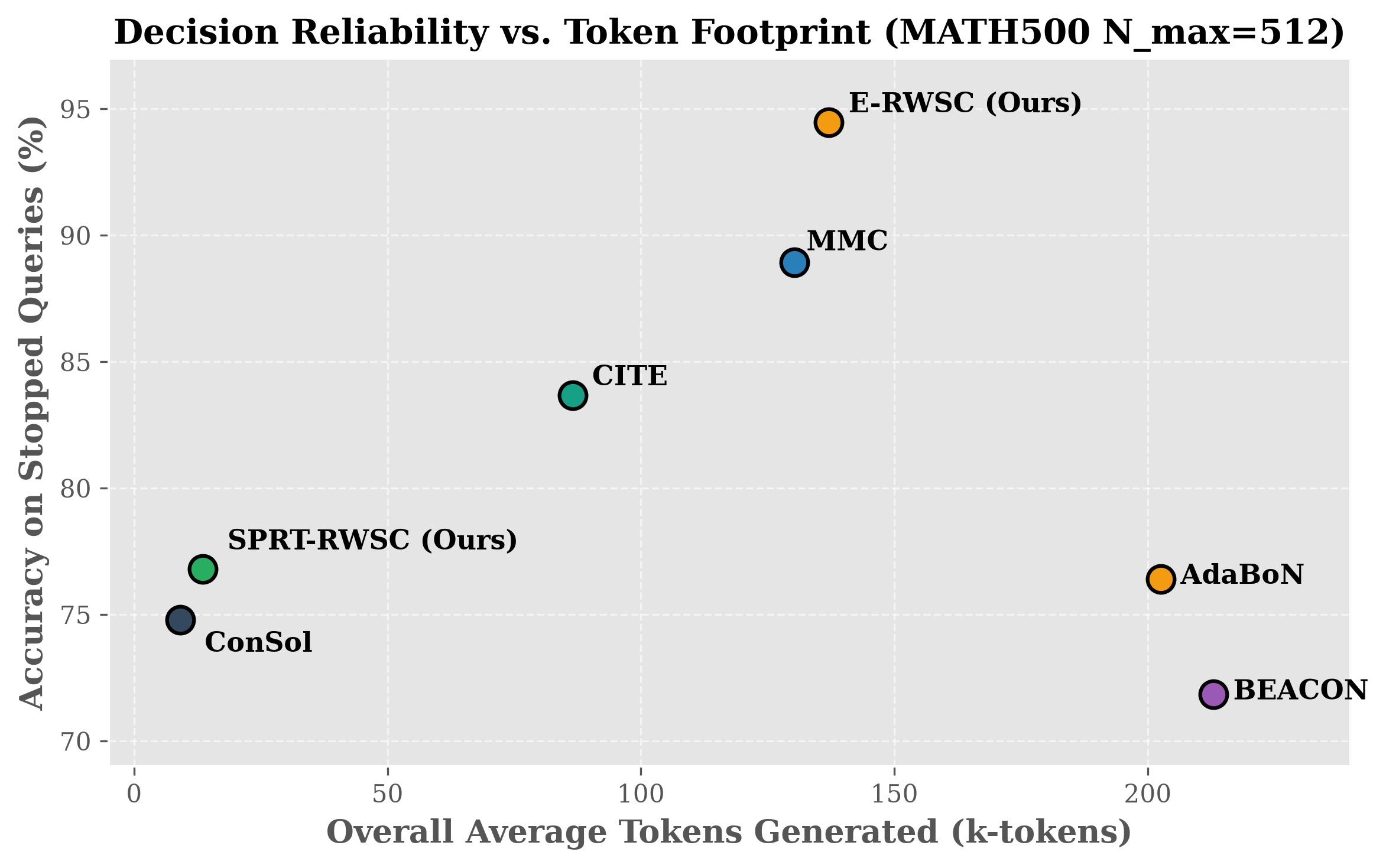}
        \caption{Decision reliability vs. overall token footprint.}
        \label{fig:tokens_vs_stopped_acc}
    \end{subfigure}
    \vspace{-2mm}
    \caption{\footnotesize{Stopping reliability and token footprint analysis on MATH500 ($N_{\max}=512$) for Qwen2.5-3B with ArmoRM. (a) \textsc{e-RWSC} achieves $94.5\%$ accuracy on certified early stops ($\tau < 512$). (b) Decision reliability on early stops plotted against overall token expenditure ($k$-tokens)}}
    \label{fig:stopping_reliability_tokens}
\end{figure}

\begin{figure}[ht]
    \centering
    \vspace*{-3mm}
    \begin{subfigure}[b]{0.32\linewidth}
        \includegraphics[width=\linewidth]{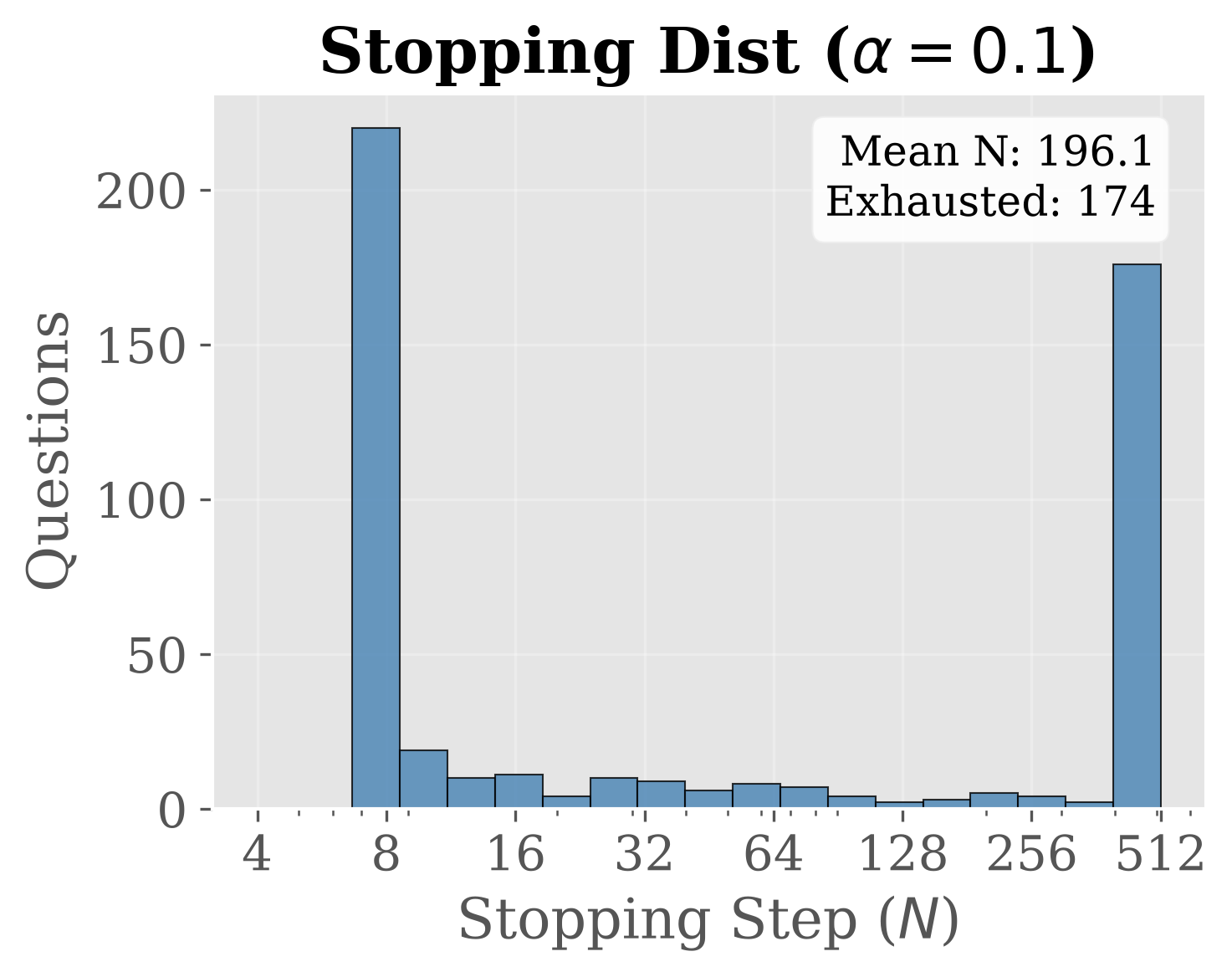}
    \end{subfigure}
    \hfill
    \begin{subfigure}[b]{0.32\linewidth}
        \includegraphics[width=\linewidth]{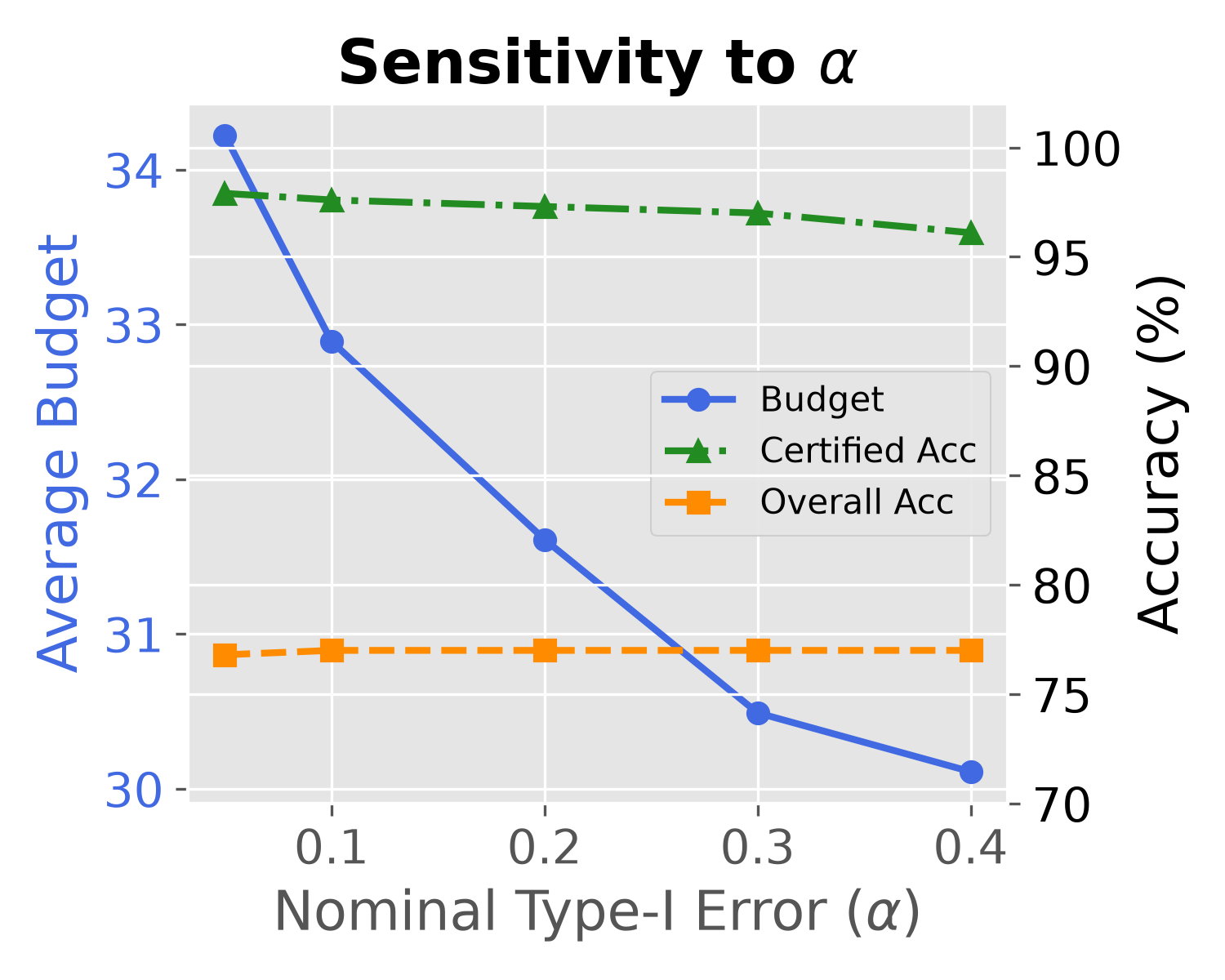}
    \end{subfigure}
    \hfill
    \begin{subfigure}[b]{0.32\linewidth}
        \includegraphics[width=\linewidth]{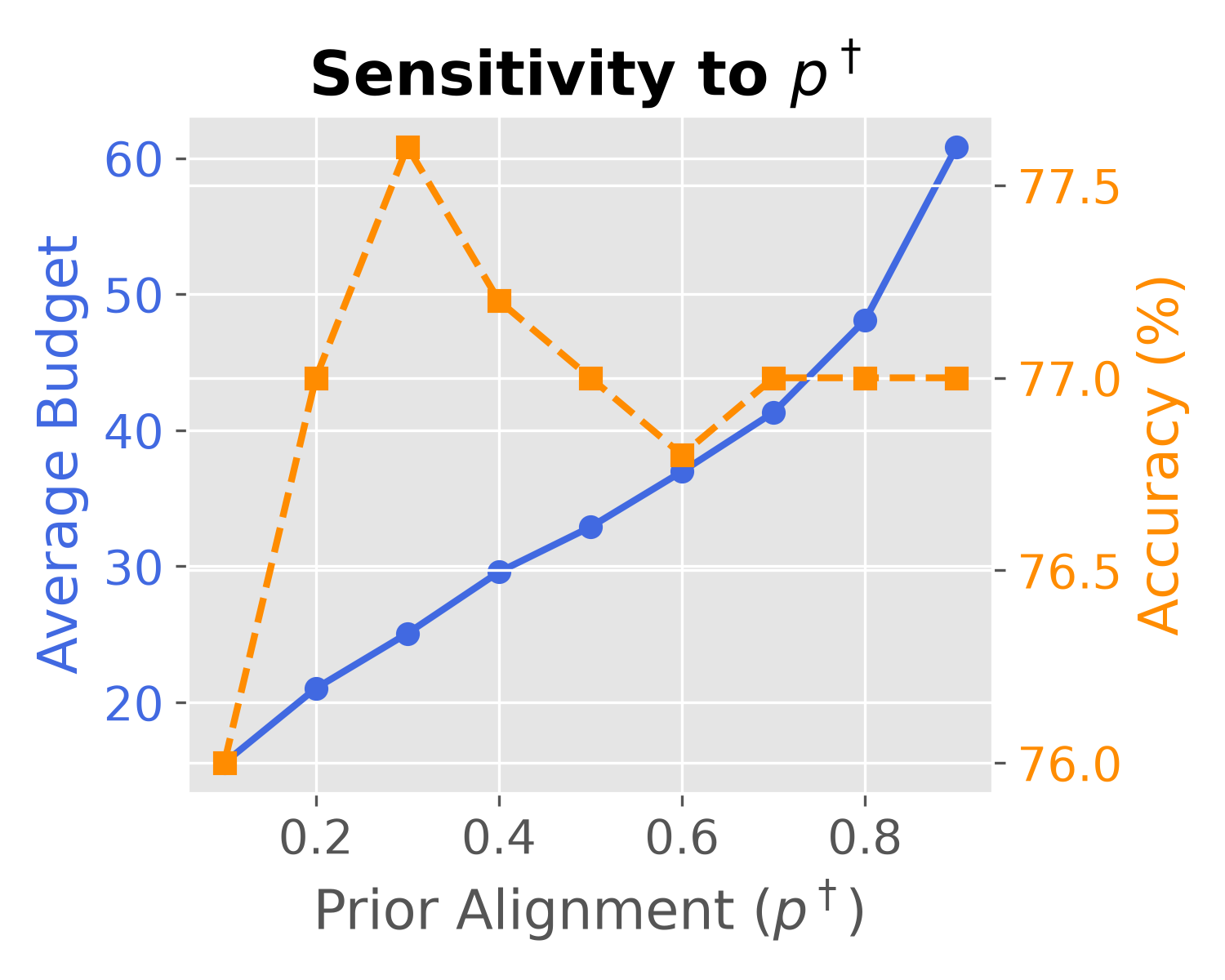}
    \end{subfigure}
    \vspace*{-4mm}
    \caption{\footnotesize{Ablation Studies on MATH500. (a) E-RWSC stopping distribution for $N_{\max}=512$. (b) Sensitivity to Type-I error boundary $\alpha$ ($N_{\max}=64$). (c) Sensitivity to the prior alignment threshold $p^\dagger$ ($N_{\max}=64$).}}
    \label{fig:ablations}
\end{figure}

\textbf{Stopping Rate and Reliability.} Figure \ref{fig:ablations}(a) shows that around 225 prompts out of 500 stop in less than 8 steps, whereas 174 extremely hard problems reach the maximum budget. At $N=512$, E-Process early stops for only $65.2\%$ of the prompts, CITE certifies $85.8\%$ of the prompts, whereas other methods like our SPRT-RWSC, Consol, and Pandora's Box BoN stop $100\%$ of the prompts early. But among these certified early-stopped answers, E-RWSC remains highly accurate. Figure \ref{fig:stopped_vs_exhausted_acc} decouples the accuracy of the certified early stopped answer from hard answers that exhausted the maximum budget. E-RWSC achieve 94.5\% accuracy on certified answers, which is more than $5\%$ accuracy gap as compared to the closest competitor in certified accuracy, MMC, with $88.9\%$ accuracy, and more than 10\% gap as compared to CITE with 83.7\% certified accuracy.

\label{sec:stopping_reliability}

\begin{table}[ht]
\centering
\resizebox{\columnwidth}{!}{%
\small
\begin{tabular}{l cc cc cc cc cc cc }
\toprule
& \multicolumn{2}{c}{\textbf{$N=24$}} & \multicolumn{2}{c}{\textbf{$N=32$}} & \multicolumn{2}{c}{\textbf{$N=64$}} & \multicolumn{2}{c}{\textbf{$N=128$}} & \multicolumn{2}{c}{\textbf{$N=256$}} & \multicolumn{2}{c}{\textbf{$N=512$}} \\
\cmidrule(lr){2-3} \cmidrule(lr){4-5} \cmidrule(lr){6-7} \cmidrule(lr){8-9} \cmidrule(lr){10-11} \cmidrule(lr){12-13}
\textbf{Method} & \textbf{Acc} & \textbf{Budget} & \textbf{Acc} & \textbf{Budget} & \textbf{Acc} & \textbf{Budget} & \textbf{Acc} & \textbf{Budget} & \textbf{Acc} & \textbf{Budget} & \textbf{Acc} & \textbf{Budget} \\
\midrule
Vanilla BoN & 71.0\% & $24.0$ & 71.4\% & $32.0$ & 69.2\% & $64.0$ & 70.2\% & $128.0$ & 69.4\% & $256.0$ & 69.8\% & $512.0$ \\
RWSC & 74.2\% & $24.0$ & 76.0\% & $32.0$ & 77.0\% & $64.0$ & 77.6\% & $128.0$ & 77.6\% & $256.0$ & 77.8\% & $512.0$ \\
Pandora & 68.6\% & $4.0$ & 68.6\% & $4.0$ & 68.6\% & $4.0$ & 68.6\% & $4.0$ & 68.6\% & $4.0$ & 68.6\% & $4.0$ \\
AdaBoN & 71.4\% & $18.8_{ \pm 5.2 }$ & 70.6\% & $24.8_{ \pm 7.3 }$ & 70.0\% & $48.5_{ \pm 15.6 }$ & 70.2\% & $95.9_{ \pm 32.2 }$ & 69.8\% & $190.8_{ \pm 65.5 }$ & 69.4\% & $380.5_{ \pm 132.0 }$ \\
BEACON & 70.8\% & $22.6_{ \pm 4.3 }$ & 71.0\% & $29.5_{ \pm 6.7 }$ & 69.0\% & $56.8_{ \pm 17.4 }$ & 70.4\% & $109.7_{ \pm 40.4 }$ & 69.6\% & $212.9_{ \pm 89.1 }$ & 70.0\% & $416.5_{ \pm 190.0 }$ \\
CONSOL & 75.4\% & $8.0_{ \pm 7.4 }$ & 74.4\% & $8.7_{ \pm 9.0 }$ & 74.6\% & $10.1_{ \pm 13.4 }$ & 74.8\% & $11.2_{ \pm 19.0 }$ & 74.8\% & $11.9_{ \pm 24.3 }$ & 74.8\% & $12.3_{ \pm 29.3 }$ \\
CITE & 75.4\% & 21.7$\pm$2.9 & 74.4\% & 26.1$\pm$6.6 & 75.6\% & 38.4$\pm$20.2 & 76.8\% & 56.5$\pm$45.9 & 77.2\% & 84.2$\pm$93.2 & 77.4\% & 124.9$\pm$175.9 \\
MMC & 75.4\% & $20.4_{ \pm 6.6 }$ & 74.4\% & $25.7_{ \pm 10.2 }$ & 75.6\% & $44.1_{ \pm 25.5 }$ & 76.8\% & $74.5_{ \pm 56.9 }$ & 77.2\% & $127.4_{ \pm 117.7 }$ & 77.4\% & $217.8_{ \pm 238.2 }$ \\
SPRT-RWSC & 74.6\% & $13.2_{ \pm 7.3 }$ & 75.8\% & $14.8_{ \pm 9.8 }$ & 76.2\% & $17.5_{ \pm 15.7 }$ & 76.4\% & $18.0_{ \pm 17.3 }$ & 76.4\% & $18.1_{ \pm 18.2 }$ & 76.4\% & $18.1_{ \pm 18.2 }$ \\
E-RWSC & 74.4\% & $15.6_{ \pm 8.2 }$ & 76.0\% & $19.3_{ \pm 12.0 }$ & 77.0\% & $32.9_{ \pm 27.1 }$ & 77.8\% & $57.9_{ \pm 57.0 }$ & 77.8\% & $105.6_{ \pm 117.1 }$ & 78.0\% & $196.1_{ \pm 236.2 }$ \\
\bottomrule
\end{tabular}%
}
\caption{MATH500 Budget Sweep Results (Qwen2.5-3B, ArmoRM)}
\label{tab:budget_sweep}
\end{table}

\textbf{Budget scaling and prevention of premature convergence.} 
Table \ref{tab:budget_sweep} illustrates the scaling behavior of selection methods as the maximum allowed budget expands from $N=24$ to $N=512$. A critical failure mode, \emph{reward hacking}, of BoN is exposed here: accuracy degrades at high budgets (dropping from $71.4\%$ at $N=32$ to $69.8\%$ at $N=512$). RWSC successfully regularizes this effect, monotonically scaling accuracy up to $77.8\%$. ConSol saturates early (halting at ${\sim}12$ for $N_{\max}=512$), resulting in stagnant accuracy even when the budget is much higher. In contrast, E-RWSC effectively bridges the gap between efficiency and inference scaling. It adaptively consumes more budget as $N_{\max}$ increases (from $15.6$ to $196.1$), selectively spending compute on highly uncertain prompts. Crucially, the early-certified subset under \textsc{e-RWSC} exhibits remarkable precision across all scales: certified accuracy stays near-ceiling, scaling from $99.2\%$ at $N=24$ (in just $8.1$ steps) to $94.5\%$ at $N=512$ ($27.5$ steps).

\textbf{Sensitivity to e-process parameters ($\alpha$ and $p^\dagger$).}
The parameters $\alpha$ (Type-I error tolerance) and $p^\dagger$ (dominance threshold) govern the anytime-valid certification guarantee. Figure~\ref{fig:ablations}(b) demonstrates that E-RWSC is robust to $\alpha$. The average budget decreases smoothly from $34.22$ to $30.11$ as $\alpha$ is relaxed from $0.05$ to $0.4$. This confirms that the practitioner can tune $\alpha$ to trade off statistical confidence for computational efficiency without a significant drop in accuracy.
Figure~\ref{fig:ablations}(c) sweeps the threshold $p^\dagger$. Because $p^\dagger$ represents the hypothesized null-agreement rate, a higher $p^\dagger$ demands stronger empirical consensus before the plug-in Kelly bet yields positive capital growth. Consequently, the generation budget scales monotonically with $p^\dagger$ (from $15.56$ at $p^\dagger=0.1$ to $60.83$ at $p^\dagger=0.9$). Accuracies peak broadly in the middle ranges ($0.3 \le p^\dagger \le 0.8$), validating our default choice of $p^\dagger=0.5$ as a sound threshold for certifying a strict majority.

\textbf{Compute Overhead.} 
To test whether sequential monitoring introduces non-trivial algorithmic latency relative to raw candidate generation, we empirically benchmark the exact wall-clock latency of every pipeline component on an NVIDIA RTX 6000 Ada GPU and AMD host CPU for MATH500 ($N_{\max}=512$). Generating each candidate response ($\sim 500$ tokens with Qwen2.5-3B) requires $7,084.5\text{ ms}$, and reward scoring with ArmoRM requires $59.8\text{ ms}$, totaling $7,144.3\text{ ms}$ of GPU compute per candidate. The CPU algorithmic overhead of \textsc{e-RWSC} for updating the e-process and evaluating the predictable Kelly bet across semantic clusters requires merely $\mathbf{13.96\ \mu\text{s}}$ ($0.014\text{ ms}$) per step. Across an entire adaptive query run ($\bar{\tau} = 196.1$), the cumulative algorithmic overhead is only $2.74\text{ ms}$, representing less than $\mathbf{0.0002\%}$ of GPU computation time ($14\times$ faster than CITE's $194.3\ \mu\text{s}$ confidence-interval check). Full hardware profiles and baseline comparisons are detailed in Appendix~\ref{app:compute_benchmark}.

\textbf{Empirical validation of theoretical results.}

\begin{wrapfigure}{r}{0.4\textwidth}
    \centering
    \vspace{-3em}
    \includegraphics[width=0.82\linewidth]{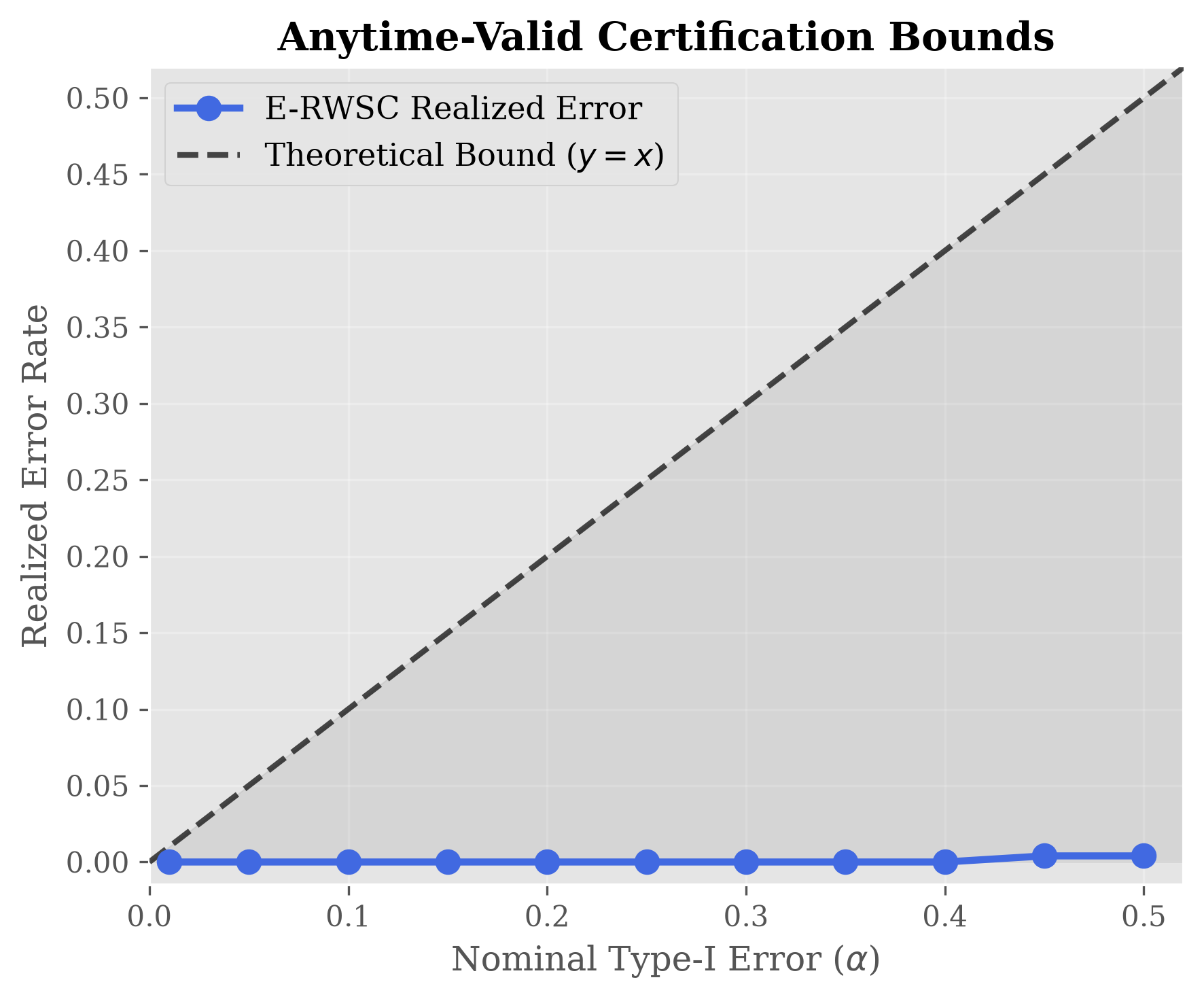}
    
    \includegraphics[width=0.82\linewidth]{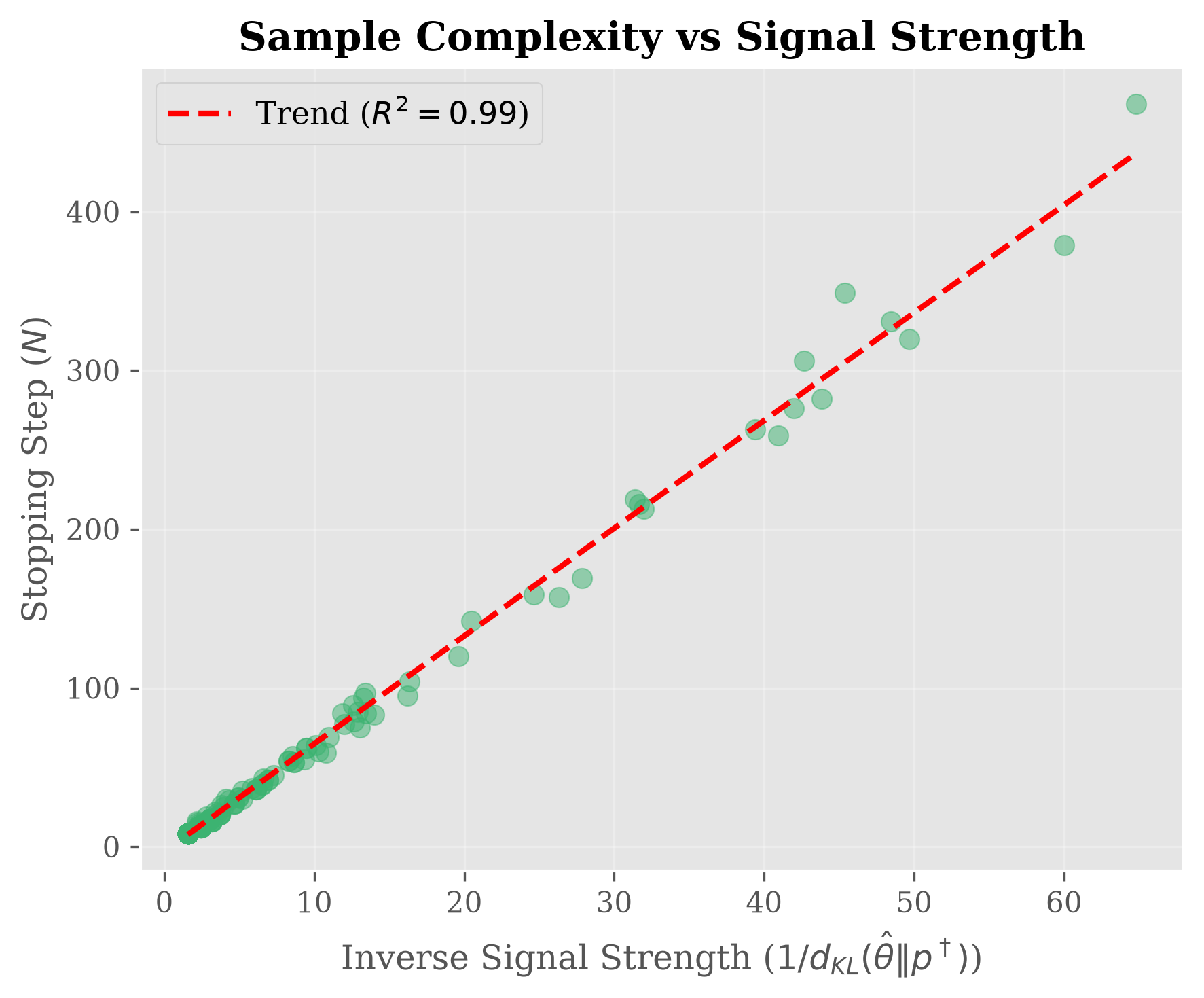}
    \vspace{-1em}
    \caption{\footnotesize{Empirical validation of E-RWSC theoretical bounds. (Top) Realized error rate respects the Type-I error ($\alpha$) threshold. (Bottom) The inverse of signal strength versus stopping step shows a linear relationship.}}
    \label{fig:eprocess_bounds}
    \vspace{-1em}
\end{wrapfigure}

We empirically validate \textsc{e-RWSC} theoretical guarantees in Figure~\ref{fig:eprocess_bounds} under dominance threshold $p^\dagger = 0.5$ (majority dominance). First, we validate the anytime-valid certification bound (Theorem~\ref{thm:av}) by sweeping the Type-I error rate $\alpha$ and measuring the realized error rate of \textsc{e-RWSC}. Following Theorem~\ref{thm:av}, an error event occurs if an answer $a^*$ is certified at stopping time $n$, but its underlying generation probability satisfies $\theta_{a^*} \le p^\dagger$; we evaluate this empirically using the candidate frequency across the full $N=512$ budget as an empirical estimate $\hat{\theta}_a$. As shown in Figure~\ref{fig:eprocess_bounds} (Top), the realized error rate remains strictly below the permitted threshold $\alpha$ across all nominal levels (empirically $\le 0.4\%$, with $0.0\%$ for all $\alpha \le 0.40$ and mean consensus $\approx 90\%$), directly validating the finite-sample anytime-valid boundary.

Second, we validate Proposition~\ref{prop:growth}, which governs the asymptotic growth rate of evidence. The theory states that the expected number of steps to stop scales with the inverse Kullback-Leibler divergence $1 / d_{\mathrm{KL}}(\hat{\theta} \| p^\dagger)$ between the empirical answer distribution and the null hypothesis. As shown in the bottom panel, there is a linear relationship between the inverse of the signal strength and the stopping time $N$, confirming that our thresholding effectively controls the early-stopping budget, which is proportional to the prompt's inherent difficulty.

\section{Conclusion}

We introduced Reward-Weighted Self-Consistency (RWSC), an answer-level aggregation rule spanning Best-of-$N$ and self-consistency, and cast the question of how many candidates a prompt needs as a sequential test over its clusters: Wald's SPRT applied to RWSC works well empirically but cannot guarantee its error rate under a shifting leader, so we introduce with \textsc{e-RWSC}, which maintains a betting e-process per answer and bounds the probability of ever certifying a non-dominant answer by $\alpha$ for \emph{any} predictable bets, at the minimax-optimal $O(\log(1/\alpha)/d_{\mathrm{KL}}(\theta_{a^\star}\|p^\dagger))$ rate under Kelly betting. Across four models, four reward models, and four benchmarks, e-RWSC matches or exceeds Best-of-$N$ accuracy at substantially reduced cost (e.g., $8.2\%$ higher accuracy than Vanilla BoN with $61.7\%$ less budget on MATH500), with realized error and per-response cost tracking these bounds empirically.  The main caveat is that certification guarantees policy dominance, $\theta_{\hat a} > p^\dagger$, not factual correctness, so a confidently wrong policy is certified just as fast as a correct one.

\bibliography{iclr2027_conference}

@article{ota2026cite,
  title={CITE: Anytime-Valid Statistical Inference in LLM Self-Consistency},
  author={Ota, Hirofumi and Iwase, Naoto and Ichihara, Yuki and Komiyama, Junpei and Imaizumi, Masaaki},
  journal={arXiv preprint arXiv:2605.05873},
  year={2026}
}

@article{cordero2025certified,
  title={Certified self-consistency: Statistical guarantees and test-time training for reliable reasoning in LLMs},
  author={Cordero-Encinar, Paula and Duncan, Andrew B},
  journal={arXiv preprint arXiv:2510.17472},
  year={2025}
}

@book{ville1939,
  title     = {{\'E}tude critique de la notion de collectif},
  author    = {Ville, Jean},
  volume={3},
  year={1939},
  publisher={Gauthier-Villars Paris}
}

@book{durrett2019probability,
  title     = {Probability: theory and examples},
  author    = {Durrett, Rick},
  volume    = {49},
  year      = {2019},
  publisher = {Cambridge university press}
}

@article{kelly1956,
  title   = {A new interpretation of information rate},
  author  = {Kelly, John L.},
  journal = {the bell system technical journal},
  volume  = {35},
  number  = {4},
  pages   = {917--926},
  year    = {1956},
  publisher={Nokia Bell Labs}
}

@article{breiman1961,
  title   = {Optimal gambling systems for favorable games},
  author  = {Breiman, L},
  journal = {The Kelly Capital Growth Investment Criterion: Theory and Practice},
  volume={3},
  pages={47},
  year={2011},
  publisher={World Scientific}
}

@article{cobbe2021gsm8k,
  title={Training verifiers to solve math word problems},
  author={Cobbe, Karl and Kosaraju, Vineet and Bavarian, Mohammad and Chen, Mark and Jun, Heewoo and Kaiser, Lukasz and Plappert, Matthias and Tworek, Jerry and Hilton, Jacob and Nakano, Reiichiro and others},
  journal={arXiv preprint arXiv:2110.14168},
  year={2021}
}

@misc{wang2023selfconsistency,
  title={Self-consistency improves chain of thought reasoning in language models},
  author={Wang, Xuezhi and Wei, Jason and Schuurmans, Dale and Le, Quoc and Chi, Ed and Narang, Sharan and Chowdhery, Aakanksha and Zhou, Denny},
  journal={arXiv preprint arXiv:2203.11171},
  year={2022}
}

@inproceedings{aggarwal2023lets,
  title={Let’s sample step by step: Adaptive-consistency for efficient reasoning and coding with llms},
  author={Aggarwal, Pranjal and Madaan, Aman and Yang, Yiming and others},
  booktitle={Proceedings of the 2023 Conference on Empirical Methods in Natural Language Processing},
  pages={12375--12396},
  year={2023}
}

@article{wald1945sequential,
author = {A. Wald},
title = {{Sequential Tests of Statistical Hypotheses}},
volume = {16},
journal = {The Annals of Mathematical Statistics},
number = {2},
publisher = {Institute of Mathematical Statistics},
pages = {117 -- 186},
year = {1945},
doi = {10.1214/aoms/1177731118},
}

@article{ramdas2023gametheoretic,
  title={Game-theoretic statistics and safe anytime-valid inference},
  author={Ramdas, Aaditya and Gr{\"u}nwald, Peter and Vovk, Vladimir and Shafer, Glenn},
  journal={Statistical Science},
  volume={38},
  number={4},
  pages={576--601},
  year={2023},
  publisher={Institute of Mathematical Statistics}
}

@inproceedings{beirami2024theoretical,
  title={Theoretical guarantees on the best-of-n alignment policy},
  author={Ahmad Beirami and Alekh Agarwal and Jonathan Berant and Alexander Nicholas D'Amour and Jacob Eisenstein and Chirag Nagpal and Ananda Theertha Suresh},
  booktitle={Forty-second International Conference on Machine Learning},
  year={2025},
  url={https://openreview.net/forum?id=u3U8qzFV7w}
}

@inproceedings{verdun2025soft,
  title={Soft Best-of-$ n $ Sampling for Model Alignment},
  author={Verdun, Claudio Mayrink and Oesterling, Alex and Lakkaraju, Himabindu and Calmon, Flavio P},
  booktitle={2025 IEEE International Symposium on Information Theory (ISIT)},
  pages={1--6},
  year={2025},
  organization={IEEE}
}

@inproceedings{coste2023reward,
  title={Reward model ensembles help mitigate overoptimization},
  author={Coste, Thomas and Anwar, Usman and Kirk, Robert and Krueger, David},
  booktitle={International Conference on Learning Representations},
  volume={2024},
  pages={50905--50931},
  year={2024}
}

@inproceedings{qiu2024treebon,
  title={TreeBoN: Enhancing Inference-Time Alignment with Speculative Tree-Search and Best-of-N Sampling.},
  author={Qiu, Jiahao and Lu, Yifu and Zeng, Yifan and Guo, Jiacheng and Geng, Jiayi and Zhu, Chenhao and Juan, Xinzhe and Yang, Ling and Wang, Huazheng and Huang, Kaixuan and others},
  booktitle={EMNLP (Findings)},
  pages={20894--20917},
  year={2025}
}

@article{chen2025sets,
  title={{SETS}: Leveraging Self-Verification and Self-Correction for Improved Test-Time Scaling},
  author={Jiefeng Chen and Jie Ren and Xinyun Chen and Chengrun Yang and Ruoxi Sun and Jinsung Yoon and Sercan O Arik},
  journal={Transactions on Machine Learning Research},
  issn={2835-8856},
  year={2025},
  url={https://openreview.net/forum?id=Wv9NMJoKww},
  note={}
}

@inproceedings{lee2025revise,
  title={ReVISE: Learning to Refine at Test-Time via Intrinsic Self-Verification},
  author={Lee, Hyunseok and Oh, Seunghyuk and Kim, Jaehyung and Shin, Jinwoo and Tack, Jihoon},
  booktitle={Proceedings of the 42nd International Conference on Machine Learning (ICML)},
  pages={33616--33634},
  volume={267},
  year={2025}
}

@article{chen2024alphamath,
  title={Alphamath almost zero: process supervision without process},
  author={Chen, Guoxin and Liao, Minpeng and Li, Chengxi and Fan, Kai},
  journal={Advances in Neural Information Processing Systems},
  volume={37},
  pages={27689--27724},
  year={2024}
}

@inproceedings{specrej2024,
title={Fast Best-of-N Decoding via Speculative Rejection},
author={Hanshi Sun and Momin Haider and Ruiqi Zhang and Huitao Yang and Jiahao Qiu and Ming Yin and Mengdi Wang and Peter Bartlett and Andrea Zanette},
booktitle={The Thirty-eighth Annual Conference on Neural Information Processing Systems},
year={2024},
url={https://openreview.net/forum?id=348hfcprUs}
}

@misc{adabon2025,
title = {AdaBoN: Adaptive Best-of-N Alignment},
author = {Vinod Raman and Hilal Asi and Satyen Kale},
year = {2026},
URL = {https://arxiv.org/abs/2505.12050}
}

@article{beacon2025,
  title={Beacon: Bayesian optimal stopping for efficient llm sampling},
  author={Wan, Guangya and Xu, Zixin Stephen and Zorc, Sasa and Baucells, Manel and Hu, Mengxuan and Wang, Hao and Li, Sheng},
  journal={arXiv preprint arXiv:2510.15945},
  year={2025}
}

@article{pandora2025,
  title={Optimal Stopping vs Best-of-$ N $ for Inference Time Optimization},
  author={Kalayci, Yusuf and Raman, Vinod and Dughmi, Shaddin},
  journal={arXiv preprint arXiv:2510.01394},
  year={2025}
}

@article{consol2025,
  title={Consol: Sequential probability ratio testing to find consistent llm reasoning paths efficiently},
  author={Lee, Jaeyeon and Qi, Guantong and Neeley, Matthew Brady and Liu, Zhandong and Jeong, Hyun-Hwan},
  journal={arXiv preprint arXiv:2503.17587},
  year={2025}
}

@inproceedings{hendrycks2021measuring,
 author = {Hendrycks, Dan and Burns, Collin and Kadavath, Saurav and Arora, Akul and Basart, Steven and Tang, Eric and Song, Dawn and Steinhardt, Jacob},
 booktitle = {Proceedings of the Neural Information Processing Systems Track on Datasets and Benchmarks},
 editor = {J. Vanschoren and S. Yeung},
 pages = {},
 title = {Measuring Mathematical Problem Solving With the MATH Dataset},
 url = {https://datasets-benchmarks-proceedings.neurips.cc/paper_files/paper/2021/file/be83ab3ecd0db773eb2dc1b0a17836a1-Paper-round2.pdf},
 volume = {1},
 year = {2021}
}

@inproceedings{hendrycks2021mmlu,
title={Measuring Massive Multitask Language Understanding},
author={Dan Hendrycks and Collin Burns and Steven Basart and Andy Zou and Mantas Mazeika and Dawn Song and Jacob Steinhardt},
booktitle={International Conference on Learning Representations},
year={2021},
url={https://openreview.net/forum?id=d7KBjmI3GmQ}
}

@inproceedings{gu2024cruxeval,
  title = 	 {{CRUXE}val: A Benchmark for Code Reasoning, Understanding and Execution},
  author =       {Gu, Alex and Roziere, Baptiste and Leather, Hugh James and Solar-Lezama, Armando and Synnaeve, Gabriel and Wang, Sida},
  booktitle = 	 {Proceedings of the 41st International Conference on Machine Learning},
  pages = 	 {16568--16621},
  year = 	 {2024},
  volume = 	 {235},
  series = 	 {Proceedings of Machine Learning Research},
  month = 	 {21--27 Jul},
  publisher =    {PMLR}
}

@inproceedings{armorm,
  title={Interpretable preferences via multi-objective reward modeling and mixture-of-experts},
  author={Wang, Haoxiang and Xiong, Wei and Xie, Tengyang and Zhao, Han and Zhang, Tong},
  booktitle={Findings of the Association for Computational Linguistics: EMNLP 2024},
  pages={10582--10592},
  year={2024}
}

@article{skywork_rm,
  title={Skywork-reward: Bag of tricks for reward modeling in llms},
  author={Liu, Chris Yuhao and Zeng, Liang and Liu, Jiacai and Yan, Rui and He, Jujie and Wang, Chaojie and Yan, Shuicheng and Liu, Yang and Zhou, Yahui},
  journal={arXiv preprint arXiv:2410.18451},
  year={2024}
}

@inproceedings{skywork-v2,
  title={Skywork-reward-v2: Scaling preference data curation via human-ai synergy},
  author={Liu, Yuhao and Zeng, Liang and Xiao, Yuzhen and He, Jujie and Liu, Jiacai and Wang, Chaojie and Yan, Rui and Shen, Wei and Zhang, Fuxiang and Xu, Jiacheng and others},
  booktitle={International Conference on Learning Representations},
  volume={2026},
  pages={133805--133838},
  year={2026}
}

@article{fsfairx,
title={{RLHF} Workflow: From Reward Modeling to Online {RLHF}},
author={Hanze Dong and Wei Xiong and Bo Pang and Haoxiang Wang and Han Zhao and Yingbo Zhou and Nan Jiang and Doyen Sahoo and Caiming Xiong and Tong Zhang},
journal={Transactions on Machine Learning Research},
issn={2835-8856},
year={2024},
url={https://openreview.net/forum?id=a13aYUU9eU},
note={}
}
\bibliographystyle{iclr2027_conference}

\newpage
\appendix

\section{Pseudocodes}\label{app:code}

\begin{algorithm}[H]
\caption{\textsc{SPRT-RWSC}}
\label{alg:sprt_rwsc}
\begin{algorithmic}[1]
\Require prompt $x$; policy $\pi_{\mathrm{ref}}$; reward model $r$; 
errors $\alpha,\beta$; budget $N_{\max}$; burn-in $n_0$
\State $A \gets \log \left(\frac{\beta}{1-\alpha}\right)$; \quad $B \gets \log \left(\frac{1-\beta}{\alpha}\right)$
\State Generate candidates $y_1, \dots, y_{n_0} \sim \pi_{\mathrm{ref}}(\cdot \mid x)$; \quad $\mathcal{A} \gets \{\mathrm{Ans}(y_i)\}_{i=1}^{n_0}$ \Comment{Burn-in phase}
\State Construct initial answer clusters and RWSC masses $S_{n_0}(x, \cdot)$
\State $a^* \gets \arg\max_{a \in \mathcal{A}} S_{n_0}(x, a)$ \Comment{Identify initial leader}
\State $\Lambda \gets 0$;\quad set $p_0,p_1$ \Comment{Initialize Log-Likelihood Ratio}
\For{$n = n_0 + 1, \dots, N_{\max}$}
    \State Sample $y_n \sim \pi_{\mathrm{ref}}(\cdot \mid x)$; \quad $A_n \gets \mathrm{Ans}(y_n)$; \quad $\mathcal{A} \gets \mathcal{A} \cup \{A_n\}$
    \State Update answer clusters and RWSC masses $S_n(x, \cdot)$ over $y_{1:n}$
    \State $a_{\mathrm{new}} \gets \arg\max_{a \in \mathcal{A}} S_n(x, a)$ \Comment{Identify current leader}
    \If{$a_{\mathrm{new}} \neq a^*$} \Comment{Leader Shift}
        \State $a^* \gets a_{\mathrm{new}}$;\quad $\Lambda \gets 0$;\quad Update $p_0,p_1$
    \Else
        \State $X \gets \mathbf{1}\{A_n = a^*\}$ \Comment{Check for agreement}
        \State $\Delta \Lambda \gets X \log \left(\frac{p_1}{p_0}\right) + (1 - X) \log \left(\frac{1 - p_1}{1 - p_0}\right)$;\quad $\Lambda \gets \Lambda + \Delta \Lambda$
    \If{$\Lambda \ge B$} \Comment{Accept $H_1$}
        \State \textbf{return} $a^*$ \Comment{Stop early and return leader}
    \ElsIf{$\Lambda \le A$} \Comment{Reject $H_1$}
        \State $\Lambda \gets 0$;\quad Update $p_0,p_1$ \Comment{Reset evidence and track new trend}
    \EndIf
    \EndIf
\EndFor
\State \textbf{return} $a^*$ \Comment{Budget exhausted, return current leader}
\end{algorithmic}
\end{algorithm}

\begin{algorithm}[H]
\caption{\textsc{e-RWSC}}
\label{alg:e_rwsc}
\begin{algorithmic}[1]
\Require prompt $x$; policy $\pi_{\mathrm{ref}}$; reward model $r$;
level $\alpha$; threshold $p^\dagger$ (default $\tfrac12$); clip $c \in (0,1)$;
prior strength $m$; budget $N_{\max}$; weights $\pi_j = \tfrac{1}{j(j+1)}$
\State $\mathcal{A} \gets \emptyset$; \; $J \gets 0$
  \Comment{active answers; count of distinct answers seen}
\For{$n = 1, 2, \dots, N_{\max}$}
    \For{\textbf{each} $a \in \mathcal{A}$}
      \Comment{bets are predictable: computed \emph{before} $y_n$ is drawn}
        \State $\hat{\theta}^{(a)} \gets
          \dfrac{Z^{(a)} + m\,S_{n-1}(x, a)}{(n - 1 - \sigma_a) + m}$
          \Comment{RWSC mass $\widehat{S}$ from Eq.~\ref{eq:rwsc_score} on $y_{1:n-1}$}
        \State $\lambda^{(a)} \gets
          \mathrm{clip}\!\Bigl(
          \tfrac{\hat{\theta}^{(a)} - p^\dagger}{p^\dagger(1 - p^\dagger)},
          \, 0, \, \tfrac{c}{p^\dagger} \Bigr)$
          \Comment{plug-in Kelly bet, Eq.~\ref{eq:plugin_bet}}
    \EndFor
    \State sample $y_n \sim \pi_{\mathrm{ref}}(\cdot \mid x)$; \;
      score $r(x, y_n)$; \;
      $A_n \gets \mathrm{Ans}(y_n)$
    \State recompute
      RWSC masses $S_n(x, \cdot)$ over $y_{1:n}$
    \For{\textbf{each} $a \in \mathcal{A}$}
      \Comment{settle all active bets against the new observation}
        \State $X \gets \mathbf{1}\{A_n = a\}$; \quad
          $E^{(a)} \gets E^{(a)} \cdot
          \bigl(1 + \lambda^{(a)}(X - p^\dagger)\bigr)$; \quad
          $Z^{(a)} \gets Z^{(a)} + X$
    \EndFor
    \If{$A_n \notin \mathcal{A}$}
      \Comment{lazy instantiation; first observation consumed, not bet on}
        \State $\mathcal{A} \gets \mathcal{A} \cup \{A_n\}$; \;
          $J \gets J + 1$; \; $J(A_n) \gets J$
        \State $\sigma_{A_n} \gets n$; \quad
          $E^{(A_n)} \gets 1$; \quad $Z^{(A_n)} \gets 0$
    \EndIf
    \If{$\exists\, a \in \mathcal{A} : \pi_{J(a)}\, E^{(a)} \ge 1/\alpha$}
      \Comment{Ville boundary, Eq.~\ref{eq:stopping_time}}
        \State $a^* \gets \arg\max_{a \in \mathcal{A}} \pi_{J(a)} E^{(a)}$
        \State \Return $y^* \gets
          \arg\max_{y \in \mathcal{Y}_{a^*}} r(x, y)$
          \Comment{\textbf{certified:} $\theta_{a^*} > p^\dagger$ w.p.\ $\ge 1-\alpha$ (Thm.~\ref{thm:av})}
    \EndIf
\EndFor
\State \Return $y^*_{\mathrm{RWSC}}$ from~\eqref{eq:rwsc_ans}
  \Comment{budget exhausted: uncertified fallback}
\end{algorithmic}
\end{algorithm}

\section{Deferred Proofs}
\label{app:proof}

\begin{proof}[Proof of Theorem~\ref{thm:av}]
Let $\mathcal{N} = \{a : \theta_a \le p^\dagger\}$ denote the set of null
answers, and for each answer $a$ let
\begin{equation}
C_a \;=\; \bigl\{ \exists\, n \ge \sigma_a : \pi_{J(a)} E^{(a)}_n \ge
1/\alpha \bigr\}
\end{equation}
be the event that $a$ is certified at some time. Note that
$C_a \subseteq \{\sigma_a < \infty\}$, since $E^{(a)}$ is instantiated only once $a$ has been observed. We must bound
$\mathbb{P}\bigl( \bigcup_{a \in \mathcal{N}} C_a \bigr)$.

Fix $a \in \mathcal{N}$ and work on the event $\{\sigma_a < \infty\}$, since outside this event there is no supermartingale — the process was never started, and $J(a)$ isn't even defined. The candidates
are i.i.d.\ draws from $\pi_{\mathrm{ref}}(\cdot \mid x)$ and $\sigma_a$ is a
stopping time, so conditionally on $\mathcal{F}_{\sigma_a}$ the post-$\sigma_a$
observations $\bigl(X^{(a)}_i\bigr)_{i > \sigma_a}$ are i.i.d.\
$\mathrm{Bernoulli}(\theta_a)$ and independent of $\mathcal{F}_{\sigma_a}$;
starting the product at $\sigma_a + 1$ is what excludes the deterministic
success $X^{(a)}_{\sigma_a} = 1$ implied by the definition of $\sigma_a$.
Each factor $U_i = 1 + \lambda^{(a)}_i (X^{(a)}_i - p^\dagger)$ is bounded
below by $1 - c > 0$, since worst case $X_i^{(a)} = 0$ gives
$1 - \lambda_i^{(a)} p^\dagger$ and $\lambda^{(a)}_i \le c/p^\dagger$, and the bets
are $\mathcal{F}_{i-1}$-measurable, so
\begin{equation}
\mathbb{E}\bigl[ U_i \,\big|\, \mathcal{F}_{i-1} \bigr]
\;=\; 1 + \lambda^{(a)}_i \bigl( \theta_a - p^\dagger \bigr) \;\le\; 1,
\end{equation}
using $\lambda^{(a)}_i \ge 0$ and $\theta_a \le p^\dagger$. Hence, under the
conditional law given $\mathcal{F}_{\sigma_a}$, the process
$n \mapsto (E_n^{(a)})_{n \ge \sigma_a}$ is a nonnegative supermartingale started
at $1$.

Ville's inequality bounds $\mathbb{P}(\sup_n M_n \ge u) \le 1/u$ for any
nonnegative supermartingale with $M_0 = 1$. The rank $J(a)$ is determined by
the answers observed up to and including time $\sigma_a$, hence is
$\mathcal{F}_{\sigma_a}$-measurable; applying Ville at threshold
$u = 1/(\pi_{J(a)}\alpha)$ gives
\begin{equation}
\label{eq:ville_cond}
\mathbb{P}\bigl( C_a \,\big|\, \mathcal{F}_{\sigma_a} \bigr)=\mathbb{P}\Bigl( \sup_{n \ge \sigma_a} E_n^{(a)} \ge
\tfrac{1}{\pi_{J(a)} \alpha} \,\Big|\, \mathcal{F}_{\sigma_a} \Bigr)
\;\le\; \pi_{J(a)}\,\alpha
\qquad \text{on } \{\sigma_a < \infty\}.
\end{equation}
Since $C_a \subseteq \{\sigma_a < \infty\}$ and
$\{\sigma_a < \infty\} \in \mathcal{F}_{\sigma_a}$, the tower property gives
\begin{align}
  \mathbb{P}(C_a)
= \mathbb{E}\bigl[ \mathbb{E}\bigl[ \mathbf{1}_{C_a}\mid \mathcal{F}_{\sigma_a}\bigr]\bigr]
= \mathbb{E}\bigl[ \mathbb{E}\bigl[ \mathbf{1}_{C_a} \mathbf{1}\{\sigma_a < \infty\} \mid \mathcal{F}_{\sigma_a}\bigr]\bigr]
&= \mathbb{E}\bigl[ \mathbb{P}( C_a \mid \mathcal{F}_{\sigma_a})\,
\mathbf{1}\{\sigma_a < \infty\} \bigr]\nonumber\\
&\le \alpha\, \mathbb{E}\bigl[ \pi_{J(a)}\,
\mathbf{1}\{\sigma_a < \infty\} \bigr]. \label{eq:unconditional}
\end{align}
The set of possible answer strings is countable, so countable subadditivity
applies. Combining with \eqref{eq:unconditional} and exchanging sum and
expectation by Tonelli's theorem (all terms nonnegative),
\begin{equation}
\label{eq:union}
\mathbb{P}\Bigl( \bigcup_{a \in \mathcal{N}} C_a \Bigr)
\;\le\; \sum_{a \in \mathcal{N}} \mathbb{P}(C_a)
\;\le\; \alpha\, \mathbb{E}\Bigl[
\sum_{a \in \mathcal{N},\, \sigma_a < \infty} \pi_{J(a)} \Bigr]
\;\le\; \alpha\, \mathbb{E}\Bigl[
\sum_{a :\, \sigma_a < \infty} \pi_{J(a)} \Bigr],
\end{equation}
the last inequality by dropping the restriction to $\mathcal{N}$, which only adds nonnegative terms.

Fix a sample path. Answers are enumerated in order of first appearance, so
$a \mapsto J(a)$ is injective on the observed answers
$\{a : \sigma_a < \infty\}$; the ranks are therefore distinct positive
integers and
\begin{equation}
\label{eq:telescope}
\sum_{a :\, \sigma_a < \infty} \pi_{J(a)}
\;\le\; \sum_{j \ge 1} \pi_j
\;=\; \sum_{j \ge 1} \Bigl( \frac{1}{j} - \frac{1}{j+1} \Bigr)
\;=\; 1 .
\end{equation}
This holds for \emph{every} sample path, not merely in expectation, so the
random variable inside the expectation in \eqref{eq:union} is bounded above
by the constant $1$; hence its expectation is at most $1$.

Substituting \eqref{eq:telescope} into \eqref{eq:union} gives
$\mathbb{P}\bigl( \bigcup_{a \in \mathcal{N}} C_a \bigr) \le \alpha$.
Truncation at $N_{\max}$ only removes elements from each $C_a$, so the bound
holds for any budget; and since Ville's inequality controls the supremum over
all times, the guarantee is unaffected by when the analyst chooses to stop,
which is the anytime-validity property.
\end{proof}

\begin{proof}[Proof of Lemma~\ref{lem:kelly}]
Obseve that $G_\theta(\lambda) = \mathbb{E}[\log(1 + \lambda(X - p^\dagger))]
= \theta\log(1 + \lambda(1-p^\dagger)) + (1-\theta)\log(1 - \lambda p^\dagger)$.
Strict concavity is immediate since $G_\theta''(\lambda) < 0$ on the
admissible interval (a sum of two negative terms). Setting
$G_\theta'(\lambda) = 0$ gives
$\theta(1-p^\dagger)/(1+\lambda(1-p^\dagger)) =
(1-\theta)p^\dagger/(1-\lambda p^\dagger)$; cross-multiplying and cancelling the terms linear in $\lambda$ yields
$\theta(1-p^\dagger) - \lambda p^\dagger(1-p^\dagger) = (1-\theta)p^\dagger$,
which gives the bet. Substituting $\lambda^\star(\theta)$ into $G_\theta$
and simplifying, using $1 + \lambda^\star(1-p^\dagger) = \theta/p^\dagger$ and
$1 - \lambda^\star p^\dagger = (1-\theta)/(1-p^\dagger)$, gives
the growth rate.
\end{proof}

\begin{proof}[Proof of Proposition~\ref{prop:growth}]

Write $\theta = \theta_{a^*}$. Throughout, ``$\to$'' denotes almost-sure convergence unless stated otherwise, and all statements are made on the event $\{\sigma_{a^\star} < \infty\}$, which has probability one since $\theta > 0$. Recall the expected one-step log-increment
$$G_\theta(\lambda) = \mathbb{E}_{X \sim \mathrm{Ber}(\theta)}
[\log(1 + \lambda(X - p^\dagger))]
= \theta \log(1 + \lambda(1-p^\dagger)) + (1-\theta)\log(1 - \lambda p^\dagger).$$
Abbreviate $\lambda_i = \lambda_i^{(a^\star)},
X_i = X_i^{(a^\star)},\xi_i = \log(1 + \lambda_i(X_i - p^\dagger))$, so
that $\log E_n^{(a^\star)} = \sum_{i=\sigma_{a^\star}+1}^{n} \xi_i$.

\emph{(i) The plug-in bets converge.} Conditionally on
$\mathcal{F}_{\sigma_{a^\star}}$, the post-instantiation indicators
$(X_i)_{i > \sigma_{a^\star}}$ are i.i.d.\ $\mathrm{Bernoulli}(\theta)$ (as in
the proof of Theorem~\ref{thm:av}). By the strong law of large
numbers, the empirical agreement
rate $\tfrac{Z_{i-1}^{(a^\star)}}{i-1-\sigma_{a^\star}} \to \theta$. The plug-in
estimate (\eqref{eq:plugin_bet}) is
$\hat\theta_{i-1}^{(a^\star)} =
\tfrac{Z_{i-1}^{(a^\star)} + m\,S_{i-1}(x,a^\star)}
{(i-1-\sigma_{a^\star}) + m}$,
a convex combination of the empirical rate and the bounded prior (RWSC mass)
$S_{i-1}(x,a^\star) \in [0,1]$. The prior weight
$\tfrac{m}{(i-1-\sigma_{a^\star}) + m} \to 0$. Hence
$\hat\theta_{i-1}^{(a^\star)} \to \theta$ regardless of the prior sequence. The Kelly map
$\theta' \mapsto \mathrm{clip}(\lambda^\star(\theta'), 0, c/p^\dagger)$ is
continuous (a clip of the continuous map $\lambda^\star$), so by the continuous
mapping theorem, the plug-in bet $\lambda_i \to \lambda^\star(\theta)$ when $\lambda^\star(\theta) \le  c/p^\dagger$.

\emph{(ii) Convergence of captial.} The bet
$\lambda_i$ is $\mathcal{F}_{i-1}$-measurable by construction, and $X_i$ is
independent of $\mathcal{F}_{i-1}$ with mean $\theta$, so
$\mathbb{E}[\xi_i \mid \mathcal{F}_{i-1}] = G_\theta(\lambda_i)$. The map
$G_\theta$ is continuous on the compact interval $[0, c/p^\dagger]$ (a sum of
logarithms with arguments bounded below by $1-c > 0$), so by continuity and Step (i), $G_\theta(\lambda_i) \to G_\theta(\lambda^\star(\theta))$. The
Ces\`{a}ro lemma~\citep{durrett2019probability} gives
\begin{equation}\label{eq:exp}
\frac{1}{n} \sum_{i=\sigma_{a^\star}+1}^{n}
\mathbb{E}[\xi_i \mid \mathcal{F}_{i-1}] \;\to\; G_\theta(\lambda^\star(\theta)).
\end{equation}
Let
$D_i = \xi_i - \mathbb{E}[\xi_i \mid \mathcal{F}_{i-1}]$, so that
$(D_i)_i$ is a martingale-difference sequence with respect to
$(\mathcal{F}_i)_i$. Because $\lambda_i \in [0, c/p^\dagger]$, each $\xi_i$ lies
in the bounded interval
$[\log(1-c),\, \log(1 + c(1-p^\dagger)/p^\dagger)]$, so $|D_i| \le L$ for a
constant $L < \infty$; in particular $\sum_i \mathrm{Var}(D_i)/i^2 \le
\sum_i L^2/i^2 < \infty$. Then, the martingale strong law
\citep{durrett2019probability} yields
$\tfrac{1}{n}\sum_{i \le n} D_i \to 0$. Adding this with \eqref{eq:exp}, and noting the finitely many terms with index at most
$\sigma_{a^\star}$ contribute $o(n)$ and so do not affect the Ces\`{a}ro
limit, gives $\tfrac{1}{n}\log E_n^{(a^\star)} \to G_\theta(\lambda^*(\theta))= d_{\mathrm{KL}}(\theta \,\|\, p^\dagger)$ whenever $\lambda^\star(\theta) \le  c/p^\dagger$.

\emph{(iii) Certification time.} Fix a sample path in the probability-one event $\{\sigma_{a^\star} < \infty\}$,
on which $\tfrac1n \log E_n^{(a^\star)} \to d_{\mathrm{KL}}(\theta \,\|\, p^\dagger)$. Since $\theta > p^\dagger$, we have $d_{\mathrm{KL}}(\theta \,\|\, p^\dagger) > 0$ and hence
$\log E_n^{(a^\star)} \to \infty$. Write
$b_\alpha = \log(1/(\pi_{J(a^\star)}\alpha))$ for the log-boundary of
$a^\star$. Note $J(a^\star)$, hence $\pi_{J(a^\star)}$, is fixed on this path
and does not depend on $\alpha$, and $b_\alpha \to \infty$ as $\alpha \to 0$.
Hence $N_{a^\star}(\alpha) = \min\{ n : \log E_n^{(a^\star)} \ge b_\alpha \}$ is
finite for every $\alpha$ since the walk diverges, and
$N_{a^\star}(\alpha) \to \infty$ as $\alpha \to 0$. Now, by definition,
\begin{equation}
\label{eq:sandwich}
\log E_{N_{a^\star}(\alpha) - 1}^{(a^\star)}
\;<\; b_\alpha \;\le\; \log E_{N_{a^\star}(\alpha)}^{(a^\star)} .
\end{equation}
On the fixed path, the sequence
$m \mapsto \tfrac1m \log E_m^{(a^\star)}$ converges to $d_{\mathrm{KL}}(\theta \,\|\, p^\dagger)$, so along
the divergent integer sequence $m = N_{a^\star}(\alpha)$ (and likewise
$m = N_{a^\star}(\alpha) - 1$, using
$\tfrac{N_{a^\star}(\alpha)-1}{N_{a^\star}(\alpha)} \to 1$) both outer terms of
\eqref{eq:sandwich}, divided by $N_{a^\star}(\alpha)$, tend to $d_{\mathrm{KL}}(\theta \,\|\, p^\dagger)$.
Hence $b_\alpha / N_{a^\star}(\alpha) \to d_{\mathrm{KL}}(\theta \,\|\, p^\dagger)$, i.e.\
$N_{a^\star}(\alpha)/b_\alpha \to d_{\mathrm{KL}}(\theta \,\|\, p^\dagger)$. Since
$b_\alpha/\log(1/\alpha) = 1 + \log(1/\pi_{J(a^\star)})/\log(1/\alpha) \to 1$,
we may replace $b_\alpha$ by $\log(1/\alpha)$, yielding
$N_{a^\star}(\alpha)/\log(1/\alpha) \to 1/d_{\mathrm{KL}}(\theta \,\|\, p^\dagger)$ a.s.

\emph{(iv) Stopping time.} $N$ of
\eqref{eq:stopping_time} halts at the first answer to cross its boundary, so
$N \le N_{a^\star}(\alpha)$ always, with equality when no other
answer is certified first. When $p^\dagger \ge \tfrac12$, $a^\star$ is the
unique answer with $\theta_a > p^\dagger$, so any other certified answer is non-dominant. By Theorem~\ref{thm:av}, this occurs with probability at most
$\alpha$. Thus $\mathbb{P}(N \neq N_{a^\star}(\alpha)) \le \alpha$. Fix
$\varepsilon > 0$. Since $N_{a^\star}(\alpha)/\log(1/\alpha) \to
1/d_{\mathrm{KL}}(\theta \,\|\, p^\dagger)$ almost surely, and hence in probability. So
$$\mathbb{P}\bigl( | N_{a^\star}(\alpha)/\log(1/\alpha) - 1/d_{\mathrm{KL}}(\theta \,\|\, p^\dagger) | >
\varepsilon \bigr) \to 0.$$
Combined with
$\mathbb{P}(N \neq N_{a^\star}(\alpha)) \le \alpha \to 0$, a union bound gives
$$\mathbb{P}\bigl( | N/\log(1/\alpha) - 1/d_{\mathrm{KL}}(\theta \,\|\, p^\dagger) | > \varepsilon \bigr)
\to 0,$$
which completes the proof.
\end{proof}

\section{Comparison with baselines}

Table~\ref{tab:qwen_armorm} reports the full breakdown underlying the summary in Section~\ref{sec:experiments}. The Oracle@24 row upper-bounds what any selection rule can achieve from this candidate pool, and the gap to it is largest on CruxEval (76.5\% vs.\ our best 54.8\%) and smallest on GSM8K (96.5\% vs.\ 91.4\%), reflecting that code-execution answers are harder to concentrate reward mass on than short numeric ones. The base policy $\pi_{\text{ref}}$ alone already reaches 81.1\% on GSM8K without any selection, so the relative gains from BoN-style methods are naturally compressed there; on MATH500, MMLU, and CruxEval, where the base policy starts much lower (63.6\%, 65.8\%, 44.0\%), selection contributes proportionally more of the final accuracy. Soft BoN ~\citep{verdun2025soft} underperforms Vanilla BoN on every benchmark, indicating that softening the arg-max selection with a temperature is not by itself sufficient to recover the consensus robustness that RWSC's explicit answer-level aggregation provides. Among the baselines, BEACON~\citep{beacon2025} takes a very high budget, where as for others, budget usage is highly Reward Model dependent or benchmark-dependent rather than a fixed property of the method: Pandora's Box BoN~\citep{pandora2025} is highly dependent on the scale of output of reward modes, with very confident answers of ArmoRM, it stops at budget same as burn-in phase, although it changes as we change the reward model, as it can be seen in Table \ref{tab:cross_rm_math500}, ConSol's budget ranges from $5.39$ on MMLU to $11.99$ on CruxEval, and CITE's stays comparatively flat ($21.60$--$23.31$) across all four, consistent with its fixed-target certification not adapting its threshold to how concentrated a given benchmark's answer distribution is. SPRT-RWSC and E-RWSC track prompt difficulty instead: both spend the most on CruxEval ($17.17$ and $20.40$, respectively) and the least on GSM8K ($11.48$ and $14.81$), mirroring the same base-policy-strength ordering that governs the oracle gap.

\begin{table}[H]
\centering
\resizebox{\textwidth}{!}{
\begin{tabular}{l cc cc cc cc}
\toprule
& \multicolumn{2}{c}{\textbf{GSM8K}} & \multicolumn{2}{c}{\textbf{MATH500}} & \multicolumn{2}{c}{\textbf{MMLU}} & \multicolumn{2}{c}{\textbf{CruxEval}} \\
\cmidrule(lr){2-3}\cmidrule(lr){4-5}\cmidrule(lr){6-7}\cmidrule(lr){8-9}
\textbf{Method} & \textbf{Acc} & \textbf{Budget} & \textbf{Acc} & \textbf{Budget} & \textbf{Acc} & \textbf{Budget} & \textbf{Acc} & \textbf{Budget} \\
\midrule
\textbf{Oracle@24 (Hard Ceiling)} & \textbf{96.5\%} & 24.00 & \textbf{90.2\%} & 24.00 & \textbf{93.4\%} & 24.00 & \textbf{76.5\%} & 24.00 \\
\midrule
\multicolumn{9}{l}{\textit{Full Budget Baselines}} \\
Base Policy ($\pi_{\text{ref}}$) & 81.1\% & 24.00 & 63.6\% & 24.00 & 65.8\% & 24.00 & 44.0\% & 24.00 \\
Vanilla BoN & 91.2\% & 24.00  & 71.0\% & 24.00  & 74.6\% & 24.00  & 48.6\% & 24.00  \\
Soft BoN & 89.1\% & 24.00 & 69.4\% & 24.00 & 70.0\% & 24.00 & 46.8\% & 24.00 \\
\midrule
\multicolumn{9}{l}{\textit{Answer-Level Aggregation}} \\
Self-Consistency & 89.8\% & 24.00 & 74.8\% & 24.00 & 70.4\% & 24.00 & 54.4\% & 24.00 \\
RWSC & 91.4\% & 24.00  & 74.2\% & 24.00  & 74.7\% & 24.00  & 54.6\% & 24.00  \\
\midrule
\multicolumn{9}{l}{\textit{Early-stopping Baselines}} \\
Pandora's Box BoN & 88.1\% & $4.00_{\pm 0.00}$ & 68.6\% & $4.00_{\pm 0.00}$ & 73.4\% & $4.00_{\pm 0.00}$ & 49.4\% & $4.00_{\pm 0.00}$ \\
AdaBoN & 91.3\% & $18.25_{\pm 5.90}$ & 71.4\% & $18.82_{\pm 5.20}$ & 74.8\% & $18.89_{\pm 5.48}$ & 48.8\% & $19.37_{\pm 5.24}$ \\
BEACON & 91.3\% & $22.06_{\pm 5.16}$ & 70.8\% & $22.55_{\pm 4.27}$ & 74.6\% & $22.26_{\pm 4.74}$ & 48.5\% & $23.16_{\pm 3.22}$ \\
ConSol & 88.0\% & $5.53_{\pm 4.81}$ & 75.4\% & $8.02_{\pm 7.38}$ & 71.8\% & $5.39_{\pm 4.56}$ & 54.1\% & $11.99_{\pm 8.50}$ \\
CITE & 87.9\% & $21.91_{ \pm 2.72 }$ & 75.4\% & $21.68_{ \pm 2.85 }$ & 72.2\% & $21.60_{ \pm 2.85 }$ & 54.4\% & $23.31_{ \pm 1.81 }$ \\
MMC & 87.9\% & $17.63_{ \pm 7.57 }$ & 75.4\% & $19.50_{ \pm 7.49 }$ & 72.2\% & $17.71_{ \pm 8.11 }$ & 54.4\% & $19.53_{ \pm 7.38 }$ \\
\midrule
\multicolumn{9}{l}{\textit{Proposed Methods}} \\
SPRT-RWSC  & 91.1\% & $11.48_{\pm 6.11}$ & 74.6\% & $13.23_{\pm 7.30}$ & 74.5\% & $12.38_{\pm 6.97}$ & 54.8\% & $17.17_{\pm 7.05}$ \\
E-RWSC     & 91.2\% & $14.81_{\pm 7.82}$ & 74.4\% & $15.59_{\pm 8.20}$ & 76.7\% & $14.19_{\pm 7.86}$ & 54.6\% & $20.40_{\pm 6.54}$ \\
\bottomrule
\end{tabular}
}
\vspace*{-2mm}
\caption{\footnotesize{Comprehensive comparison of all selection methods across GSM8K, MATH500, MMLU and CruxEval for \textbf{Qwen2.5-3B-Instruct} using \textbf{ArmoRM} reward model, with allowed maximum budget as 24. For E-Process methods, we use $\alpha=0.1$, $p^\dagger=0.5$, and $c=0.8$. For SPRT we use $\alpha{=}0.1, \beta{=}0.2$.}}
\label{tab:qwen_armorm}
\end{table}

\section{Computational Overhead}
\label{app:compute_benchmark}

A common concern in sequential early-stopping mechanisms is whether the continuous algorithmic overhead of evaluating statistical bounds at each step negates the savings gained from generating fewer tokens. To rigorously quantify this trade-off, we conduct comprehensive wall-clock hardware benchmarks on the MATH500 dataset under $N_{\max}=512$ using the \textbf{Qwen2.5-3B-Instruct} policy and \textbf{ArmoRM-Llama3-8B-v0.1} reward model. All measurements are executed on dedicated hardware: an NVIDIA RTX 6000 Ada Generation GPU (48GB VRAM) for model inference and an AMD host processor for sequential decision logic.

\paragraph{Per-Candidate Latency Decomposition.}
Table~\ref{tab:hardware_latency_breakdown} breaks down the exact execution latency per candidate response. On MATH500, the average candidate trajectory generated by Qwen2.5-3B contains $500.7$ tokens. Autoregressive LLM decoding takes $14.15\text{ ms}$ per token, resulting in an average generation latency of $7,084.50\text{ ms}$ (${\approx}7.08\text{ s}$) per candidate. A batched forward pass of ArmoRM requires an additional $59.76\text{ ms}$ to evaluate the candidate's scalar reward. Answer extraction via canonical regex parsing adds approximately $0.20\text{ ms}$ on CPU. Thus, generating and evaluating a single candidate takes \textbf{$7,144.26\text{ ms}$} (${\approx}7.14\text{ s}$) of total GPU compute. 

In sharp contrast, the sequential stopping check of \textsc{e-RWSC} requires computing predictable Kelly bets across the set of distinct semantic answer clusters $\mathcal{A}$ and updating scalar e-processes. Because empirical answer clustering produces at most $|\mathcal{A}| \le 5$ distinct answers per prompt, this step consists of simple scalar arithmetic with time complexity $\mathcal{O}(|\mathcal{A}|)$, executing in just $\mathbf{13.96\ \mu\text{s}}$ ($0.014\text{ ms}$) on CPU. The ratio of algorithmic decision overhead to raw GPU candidate compute is approximately $\mathbf{0.00020\%}$ ($\approx 1$ part in $500,000$).

\begin{table}[H]
\centering
\small
\caption{\footnotesize{\textbf{Hardware Latency Breakdown per Candidate} on MATH500 using Qwen2.5-3B-Instruct (avg $500.7$ tokens/candidate) and ArmoRM-8B on an NVIDIA RTX 6000 Ada GPU and AMD host CPU.}}
\label{tab:hardware_latency_breakdown}
\vspace{2mm}
\resizebox{\textwidth}{!}{
\begin{tabular}{llcc}
\toprule
\textbf{Pipeline Component} & \textbf{Hardware Device} & \textbf{Complexity} & \textbf{Latency / Candidate} \\
\midrule
LLM Candidate Generation (Qwen2.5-3B) & GPU (RTX 6000 Ada) & $\mathcal{O}(L \cdot d_{\text{model}})$ & $7,084.50\text{ ms}$ (${\approx}7.08\text{ s}$) \\
RM Scoring (ArmoRM-8B forward pass) & GPU (RTX 6000 Ada) & $\mathcal{O}(L \cdot d_{\text{model}})$ & $59.76\text{ ms}$ \\
Answer Extraction \& Normalization & CPU (Host) & $\mathcal{O}(1)$ regex & $\approx 0.20\text{ ms}$ \\
\textsc{e-RWSC} E-Process Update \& Kelly Bet & CPU (Host) & $\mathcal{O}(|\mathcal{A}|)$ scalar & $\mathbf{0.014\text{ ms}}$ ($13.96\ \mu\text{s}$) \\
\midrule
\textbf{Total GPU Compute per Candidate} & GPU & --- & $\mathbf{7,144.26\text{ ms}}$ (${\approx}7.14\text{ s}$) \\
\textbf{Algorithmic Overhead Ratio} & Host CPU / GPU & --- & $\mathbf{\approx 0.00020\%}$ \\
\bottomrule
\end{tabular}
}
\end{table}

\begin{table}[H]
\centering
\small
\caption{\footnotesize{\textbf{End-to-End Computational Profile and Algorithmic Overhead on MATH500 ($N_{\max}=512$)} for Qwen2.5-3B-Instruct with ArmoRM-8B. GPU generation and RM scoring latency measured on an NVIDIA RTX 6000 Ada GPU; algorithmic stopping-check overhead measured on the host CPU. Algorithmic overhead is under $0.0002\%$ of generation time for \textsc{e-RWSC}, while saving $61.7\%$ of total GPU compute time relative to fixed-budget BoN.}}
\label{tab:computational_overhead}
\vspace{2mm}
\resizebox{\textwidth}{!}{
\begin{tabular}{l ccc ccc cc}
\toprule
\textbf{Method} & \textbf{Avg Steps} & \textbf{Generated} & \textbf{GPU Compute} & \textbf{Algorithmic Check} & \textbf{Algorithmic} & \textbf{Overall} & \textbf{Certified} \\
& $\bar{N}$ & \textbf{Tokens} ($k$) & \textbf{Time} (s / query) & \textbf{Latency} ($\mu$s / step) & \textbf{Overhead} (ms / query) & \textbf{Acc (\%)} & \textbf{Acc (\%)} \\
\midrule
\multicolumn{8}{l}{\textit{Fixed-Budget Baselines}} \\
Vanilla BoN & 512.0 & 256.3 & 3657.7 (${\approx}61.0$ min) & --- & 0.000 & 69.8\% & --- \\
Self-Consistency & 512.0 & 256.3 & 3657.7 (${\approx}61.0$ min) & --- & 0.000 & 77.4\% & --- \\
\midrule
\multicolumn{8}{l}{\textit{Adaptive Sequential Baselines}} \\
AdaBoN & 380.4 & 202.6 & 2717.9 (${\approx}45.3$ min) & 0.02 $\mu$s & 0.008 & 69.4\% & 76.4\% \\
BEACON & 416.5 & 213.0 & 2975.5 (${\approx}49.6$ min) & 1.75 $\mu$s & 0.729 & 70.0\% & 71.8\% \\
Pandora's Box & 4.0 & 2.0 & 28.6 (${\approx}0.5$ min) & 0.17 $\mu$s & 0.001 & 68.6\% & 68.6\% \\
ConSol & 12.3 & 9.1 & 88.1 (${\approx}1.5$ min) & 0.13 $\mu$s & 0.002  & 74.8\% & 74.8\% \\
CITE & 124.9 & 86.5 & 892.2 (${\approx}14.9$ min) & 194.33 $\mu$s & 24.269 & 77.4\% & 83.7\% \\
\midrule
\multicolumn{8}{l}{\textit{Proposed Frameworks}} \\
SPRT-RWSC (Ours) & 21.8 & 13.6 & 155.8 (${\approx}2.6$ min) & 5.83 $\mu$s & 0.127 & 76.8\% & 76.8\% \\
\textsc{e-RWSC} (Ours) & 196.1 & 137.1 & 1400.8 (${\approx}23.3$ min) & 13.96 $\mu$s & 2.737 & \textbf{78.0\%} & \textbf{94.5\%} \\
\bottomrule
\end{tabular}
}
\end{table}

\paragraph{End-to-End Computational Profile Across Methods.}
Table~\ref{tab:computational_overhead} reports the full end-to-end computational expenditure across all baseline methods on MATH500 under $N_{\max}=512$. Key empirical takeaways include:
\begin{enumerate}
    \item \textbf{Massive GPU Savings without Latency Overhead:} Fixed-budget methods (Vanilla BoN and Self-Consistency) consume $3,657.7\text{ s}$ (${\approx}61.0\text{ minutes}$) of dedicated GPU compute per query across the 512 candidates. By halting early on confident prompts ($\bar{\tau} = 196.1$), \textsc{e-RWSC} slashes average GPU compute to $1,400.8\text{ s}$ (${\approx}23.3\text{ minutes}$)---saving \textbf{$2,256.9\text{ seconds}$ ($37.6\text{ minutes}$) of GPU runtime per query} ($61.7\%$ relative compute reduction). Over an entire query, the cumulative CPU stopping overhead of \textsc{e-RWSC} is just $2.737\text{ ms}$, which is virtually unmeasurable in wall-clock user experience.
    \item \textbf{Efficiency Superiority over CITE:} While CITE also performs certified sequential stopping, its stopping criterion requires inverting empirical-Bernstein confidence intervals over dynamic subsets of candidates at every step. This inflates CITE's per-step check latency to $194.33\ \mu\text{s}$ ($14\times$ slower than \textsc{e-RWSC}), accumulating $24.269\text{ ms}$ of CPU compute per query. Despite this higher computational cost, CITE achieves only $83.7\%$ certified accuracy, whereas \textsc{e-RWSC} achieves \textbf{$94.5\%$ certified accuracy} with negligible algorithmic overhead.
    \item \textbf{Reward Hacking Regularization:} Vanilla BoN suffers severe performance degradation at $N=512$, dropping to $69.8\%$ accuracy despite consuming over an hour of GPU compute. \textsc{e-RWSC} reaches $\mathbf{78.0\%}$ overall accuracy (and $\mathbf{94.5\%}$ on certified queries) while spending less than half the GPU compute, proving that anytime-valid stopping simultaneously optimizes computational efficiency and mathematical accuracy.
\end{enumerate}

\end{document}